\pdfoutput=1
\documentclass[]{TEAI}

\usepackage{array}
\usepackage{tabularx}
\usepackage{ragged2e}
\usepackage{float}
\usepackage{enumitem}
\usepackage{makecell}
\usepackage{nicefrac}
\usepackage{xcolor}
\usepackage[export]{adjustbox}

\setcitestyle{authoryear}

\DeclareMathOperator*{\argmin}{arg\,min}

\newcolumntype{L}[1]{>{\RaggedRight\arraybackslash}p{#1}}
\newcolumntype{Y}{>{\RaggedRight\arraybackslash}X}
\newcolumntype{C}[1]{>{\Centering\arraybackslash}p{#1}}

\newcommand{\panelnote}[1]{%
  \vspace{0.25ex}%
  \par\noindent
  {\scriptsize\RaggedRight #1\par}%
}

\graphicspath{{figures/}}

\newlength{\twopanel}
\newcommand{\udf}{\textsc{UDF}}
\newcommand{\base}{\mathrm{Base}}
\newcommand{\rl}{\mathrm{RL}}

\newcommand{\budgetset}{\mathcal{B}}
\newcommand{\policyspace}{\Pi_{\udf}}

\title{From Base Rollouts to RL Reasoning: A Budgeted Search Perspective}

\author{
    Wenhe Sun\textsuperscript{1,$*$},
    Cunxiang Wang\textsuperscript{2,3,$*$,$\dagger$},
    Zijun Yao\textsuperscript{3},
    Yixin Cao\textsuperscript{1,4,$\dagger$}
}

\affiliation[1]{\mbox{Fudan University}}
\affiliation[2]{\mbox{Zhipu AI}}
\affiliation[3]{\mbox{Tsinghua University}}
\affiliation[4]{\mbox{Shanghai Innovation Institute}}

\contribution[*]{Equal contribution}
\contribution[\dagger]{Corresponding authors}

\abstract{
Reinforcement learning with verifiable rewards (RLVR) improves language-model reasoning, but how these gains relate to inference-time decoding and search remains unclear.
Does RL produce reasoning behavior the base model does not have, or does it shift the rollout distribution toward trajectories the base model can already reach but rarely samples?
We study this behaviorally with a Unified Decoding Framework (\udf{}), which expresses token-level sampling, beam-like search, tree search, and sequence-level resampling as executable policies over a shared budgeted operating space.
Operating points are scored after generation with pass@$k$, self-consistency, best-of-$N$, and first-finish success, so the generation policy stays separate from the evaluation metric.

Using paired Base/RL checkpoints from SimpleRL-Zoo, we ask whether an RL default-policy curve can be approximated by a structured path of Base operating points $(\pi, N_{\mathrm{Base}})$ rather than by post-hoc per-budget matching.
On Math500, AIME, GPQA, and IFEval, the pass@$k$ recovery path follows a Budgeted Operating-Point Transition Rule (BOPTR), $N_{\mathrm{Base}} \approx \alpha N_{\mathrm{RL}}^{\beta}$, with benchmark-conditioned exponents.
On Qwen2.5-7B, BOPTR gives the lowest transfer error among the non-oracle rules we test, 3.41 pp (95\% CI [2.32, 5.53]); a three-seed replication gives 3.07 $\pm$ 0.39 pp.
The rule extends to ten models across four families, at 3.28 to 4.87 pp on the checkpoints added after fitting, and to four benchmarks it was never fitted on, at 5.03 pp against 4.44 pp in fit.
It also holds at 4.19 pp without an RL checkpoint for the target model, and at 5.08 pp without RL supervision of any kind.

These results support a qualified internalized-search reading: under the recipe we test, much of the measured RL gain corresponds to a change in sampling efficiency, toward operating points the base model can already reach under search.
We treat the benchmark-wise scaling patterns as descriptive of this recipe and cohort, report the cases in which they break down, and use \udf{} and BOPTR as behavioral diagnostics rather than as evidence of parameter-level equivalence or as general predictors of post-RL performance.
}

\correspondence{\email{yxcao@fudan.edu.cn}, \email{wangcunxiang303@gmail.com}}
\checkdata[Code]{\url{https://github.com/HALIS-sh/Searchlens_boptr}}

\begin{document}
\maketitle

\vspace{-1.0em}

\section{Introduction}

Reinforcement learning with verifiable rewards (RLVR) has become a central post-training recipe for improving reasoning in large language models \citep{openai2024reasoning, guo2025deepseekr1}.
Open recipes such as DeepSeekMath and SimpleRL-Zoo further show that simple verifiable-reward training can substantially improve mathematical reasoning in open base models \citep{shao2024deepseekmath, zeng2025simplerlzoo}.
Yet recent analyzes suggest that these gains may not always reflect entirely new reasoning behavior.
Base models can sometimes approach RL-level performance under larger sampling budgets, whereas RL-trained models often dominate in low-budget regimes \citep{yue2025rlvr}.
This suggests an alternative behavioral interpretation: RL may improve \emph{sampling efficiency} by making correct reasoning trajectories easier to sample under a fixed inference budget.
Related work further suggests that RL post-training can amplify behaviors already present in the pretraining distribution \citep{zhao2025echo}.
These observations motivate our central question: how does RL change the observable behavior of the base model's default rollout distribution?

% We view decoding for a single problem as allocating a finite rollout budget over an implicit reasoning space, as sketched in Fig.~\ref{fig:main-pipeline}(1).
% A base model may assign non-negligible probability mass to many trajectories, including correct ones that are difficult to sample.
% RL may shift the default rollout distribution toward trajectories that are more likely to receive reward, while external decoding and search policies may alter the same allocation by sampling more broadly, concentrating mass, or exploring alternatives.
% We use this view as a behavioral abstraction, not as a claim about the model's internal computation.
% Under this abstraction, we test a behavioral version of the internalized-search hypothesis: RL may make the model's default rollout behavior resemble a policy that external inference-time search would otherwise need to implement.

\begin{figure}[t]
    \centering
    \includegraphics[
        width=0.94\textwidth,
        trim=6 0 6 0,
        clip
    ]{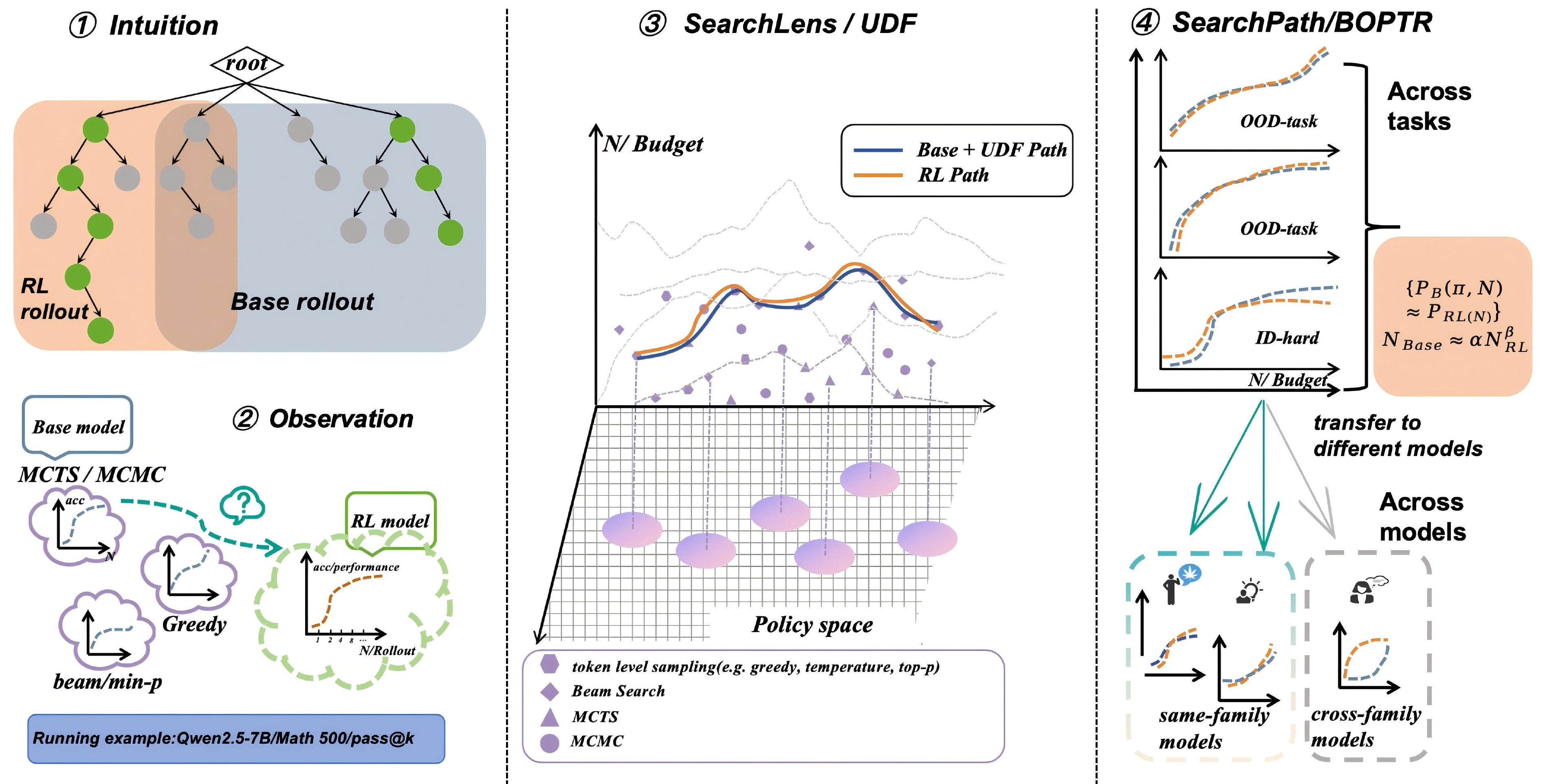}
    \caption{
    Overview of the analysis pipeline.
    We first view decoding as rollout-budget allocation over an implicit reasoning tree, then compare RL behavior with external decoding/search policies applied to the base model.
    SearchLens/\udf{} places these policies in a shared budgeted operating landscape, and SearchPath/BOPTR asks whether RL default-policy curves can be recovered as structured Base+\udf{} paths following \(N_{\text{Base}}\approx \alpha N_{\text{RL}}^{\beta}\) across tasks and models.
    }
    \label{fig:main-pipeline}
\end{figure}

We view decoding on a single problem as allocating a finite rollout budget over an implicit rollout space: the reasoning trajectories induced by the prompt and model, as illustrated in Fig.~\ref{fig:main-pipeline}(1).
A base model may spread this budget across diverse trajectories, including correct solutions in its support that are hard to sample by default.
RL can be understood as shifting the default allocation toward reward-aligned trajectories, increasing the chance of reaching successful reasoning paths under small budgets.
This shift, however, need not be unique to RL-trained parameters: external decoding and search over the base model can also reallocate budget by broadening exploration, concentrating probability mass, or selecting among alternatives.
We therefore test a behavioral form of the internalized-search hypothesis: whether the RL model's default rollout behavior can be recovered by budgeted external search over the base model.

Testing this hypothesis requires addressing three challenges.
First, decoding and search methods are heterogeneous: token-level sampling, beam-like expansion, tree-style exploration, and sequence-level resampling have different control structures.
Second, recovery can be overstated: a large policy grid may yield post-hoc matches to an RL curve, so meaningful recovery should form a low-complexity path rather than unrelated per-budget choices.
Third, such structure may depend on task regime and model family, so a rule observed on Math500 cannot be assumed to transfer unchanged to AIME, GPQA, IFEval, or models trained under different difficulty splits.

We make this question testable with \textbf{SearchLens}, a Unified Decoding Framework (\udf{}).
\udf{} represents token-level sampling, beam-like expansion, tree-style exploration, and sequence-level resampling as executable policies in a shared budgeted operating space.
Each inference-time method is paired with a budget to form an operating point \((\pi,N_{\text{Base}})\), while metrics such as pass@\(k\), self-consistency, best-of-\(N\), and first-finish success are applied only after generation.
This separation is important: we do not ask only whether Base matches RL under the same policy and budget, but whether the RL default-policy curve can be recovered somewhere in the Base model's broader policy--budget landscape.
We then ask whether the recovering points form a low-complexity path across budgets, benchmarks, and models, rather than a collection of unrelated post-hoc matches.
Such paths identify which parts of the RL gain are already accessible through Base inference-time search, while exposing regions outside the tested operating landscape.

Across paired Base/RL checkpoints from SimpleRL-Zoo, we find that RL default-policy curves are often recoverable as Base operating paths.
In the pass@\(k\) channel, the recovered Base budget follows a regime-conditioned transition rule,
\(N_{\text{Base}}\approx \alpha N_{\text{RL}}^{\beta}\):
Math500 is sublinear, GPQA and IFEval are near-linear OOD regimes, and AIME is floor-bound.
Across models, the same formula family remains informative, but the multiplier and selected policy are model-specific.
This yields a qualified internalized-search interpretation: RL often moves the model through the base operating landscape, but the concrete recovery path depends on both benchmark regime and model family.
In particular, models trained under the same SimpleRL-Zoo difficulty split share more transferable operating-point structure, whereas OOD benchmarks, cross-family models, and different training distributions mark the limits of direct transfer.

Our contributions are:
\begin{itemize}[leftmargin=*]
    \item \udf{} (SearchLens), a unified budgeted operating space for external decoding and search policies, separated from evaluation metrics.
    \item A two-dimensional recovery analysis that asks whether RL default-policy curves can be matched by Base operating points \((\pi,N_{\text{Base}})\), rather than by same-budget comparisons alone.
    \item \textbf{BOPTR-P1} (SearchPath), a low-dimensional pass@$k$ rule in which the effective Base budget follows a benchmark-regime-conditioned exponent.
    \item Horizontal and vertical transfer analyses under SimpleRL-Zoo, showing shared operating-point structure within the same difficulty split and clear limits under OOD, cross-family, or different-training-distribution settings.
\end{itemize}

Code, configs, and analysis scripts are released at \url{https://github.com/HALIS-sh/Searchlens_boptr}.

\section{Related Work}

\paragraph{RL and reward-based post-training.}
Reinforcement learning from human or verifiable rewards has become a standard approach for shaping pretrained language models.
Early instruction-following work uses RLHF with PPO over human-preference reward models \citep{ouyang2022instructgpt, schulman2017ppo}, while DPO reformulates preference optimization as a direct objective without an explicit reward model \citep{rafailov2023dpo}.
For reasoning, RL systems such as OpenAI o1 and DeepSeek-R1 demonstrate that reinforcement learning can substantially improve complex reasoning behavior \citep{openai2024reasoning, guo2025deepseekr1}.
Open RL-with-verifiable-rewards (RLVR) recipes such as DeepSeekMath and SimpleRL-Zoo use programmatic or rule-based correctness signals to train open base models \citep{shao2024deepseekmath, zeng2025simplerlzoo}.
SimpleRL-Zoo is central to our study because it provides paired Base/RL checkpoints across model scales and families under a shared verifiable-reward recipe \citep{zeng2025simplerlzoo}.
We use these checkpoints not to propose another RL method, but to analyze how RL-trained models relate behaviorally to their base counterparts under decoding and search.

\paragraph{What does RL change?}
Recent work asks whether RLVR creates new reasoning abilities or primarily reweights behaviors already present in the base distribution.
RLVR-trained models often outperform base models at small sampling budgets, while base models can recover much of the gap under large pass@\(k\), suggesting improved sampling efficiency and possible narrowing of the reasoning boundary \citep{yue2025rlvr}.
Other studies similarly frame RL gains as amplification of pretraining patterns or distribution-level generalization beyond supervised fine-tuning \citep{zhao2025echo, chu2025sft}.
\citet{karan2025sampling} further show that training-free inference-time sampling from base models can approach or outperform RL-posttrained models on several reasoning benchmarks, suggesting that some RL gains can be elicited without additional training.
This comparison is only well posed once inference budget is treated as an explicit axis: \citet{snell2024scaling} and \citet{wu2025inference} show that how a fixed test-time budget is allocated can matter as much as model scale, so a Base--RL comparison at one budget need not hold at another.
The opposite direction also has support: \citet{liu2025prorl} report that prolonged RL training expands the reasoning boundary on tasks the base model does not solve even at large sampling budgets, which confines a reweighting-only reading to the training durations and recipes actually tested.
Together, these findings motivate a behavioral view of RL: post-training may shift the default rollout distribution toward high-reward trajectories that are already accessible to the base model, with the tested recipe and budget range part of the claim rather than background conditions.
We therefore compare RL not only with the base model under its default policy, but also with the base model's full budgeted decoding and search landscape.

\paragraph{Inference-time search, selection, and behavioral analysis.}
Inference-time compute improves reasoning through chain-of-thought prompting, self-consistency, repeated sampling, explicit tree/graph/planner-style search, and verifier- or reward-model-based selection \citep{wei2022cot, wang2023selfconsistency, brown2024monkeys, yao2023tot, besta2024got, hao2023rap, zhou2024lats, cobbe2021gsm8k, lightman2023verify, chow2025bon}.
Our \udf{} framework represents these generation and search mechanisms as budgeted policies in a common operating space, while metrics such as pass@\(k\), self-consistency, best-of-\(N\), and first-finish success summarize the resulting candidate sets.
Because chain-of-thought rationales need not faithfully describe a model's internal computation \citep{turpin2023unfaithful, lanham2023faithfulness, jacovi2020faithfulness}, we focus on observable input--output behavior rather than parameter-level mechanisms.
This positions our analysis between parameter-space studies of post-training and purely outcome-level benchmark comparisons.

\section{SearchLens: A Unified Decoding/Search Landscape}
\label{sec:udf}

A decoding or search method is an inference-time policy that maps a prompt to candidate responses.
SearchLens, our Unified Decoding Framework (\udf{}), represents each method as an operating point
\begin{equation}
    u=(\pi,n),
\end{equation}
where \(\pi\in\policyspace\) is a policy configuration and \(n\in\budgetset\) is an inference budget.
Evaluation metrics are applied only after generation and are not part of the policy definition.
In our experiments, \(\budgetset=\{1,2,4,8,16\}\).
Each operating point specifies one allocation of a finite rollout budget over possible trajectories, without assuming that this space is internally represented by the base model.

\begin{figure}[t]
    \centering
    \includegraphics[width=0.72\linewidth]{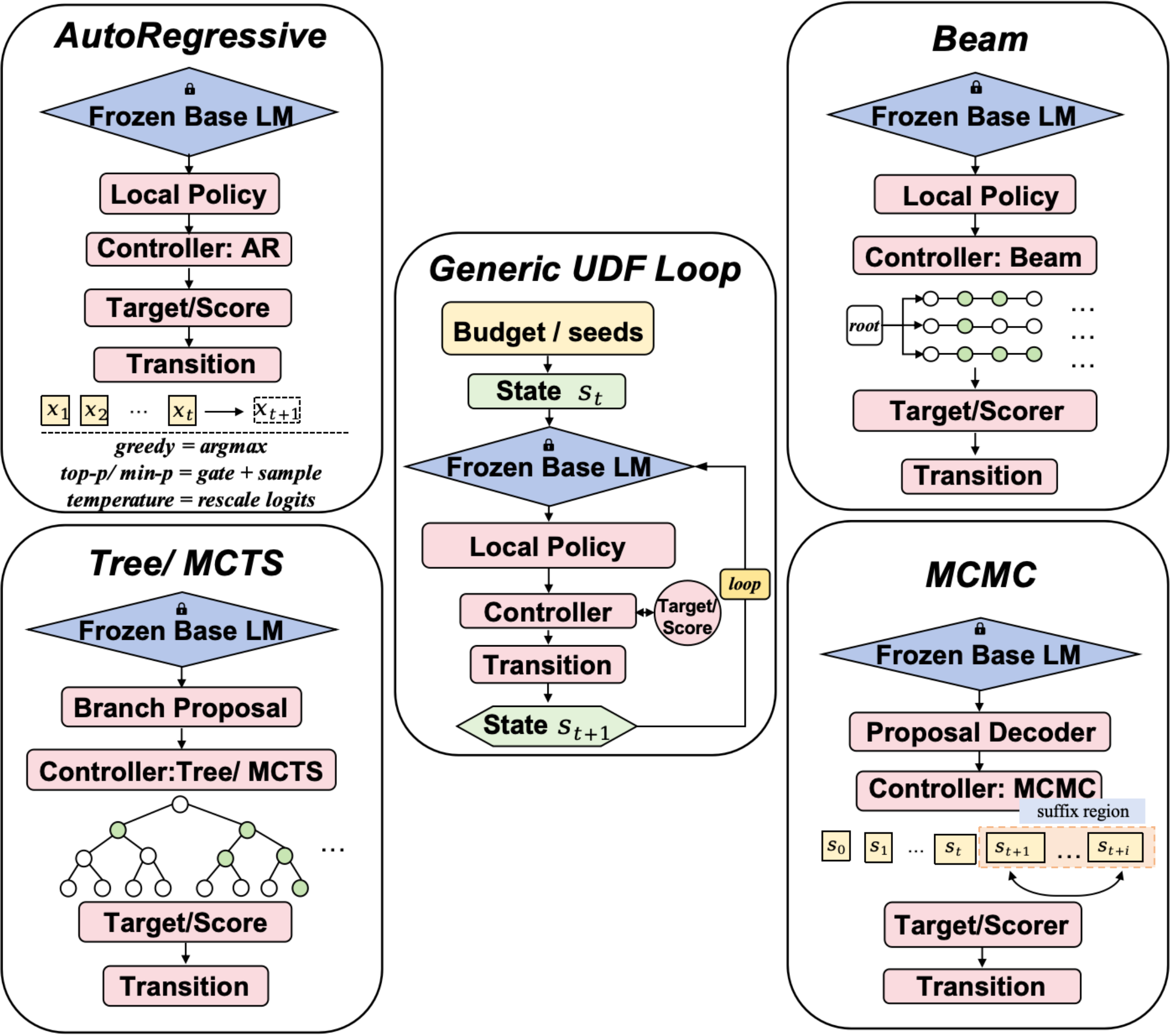}
    \caption{The Unified Decoding Framework (\udf{}). A frozen base language model is wrapped by a five-component policy---local policy, controller, evaluator, transition, and scheduler---which together define an operating point \(u=(\pi,n)\) in the budgeted landscape. Evaluation metrics, including pass@\(k\), SC, BoN, and FFS, are applied after generation and are not part of \udf{}.}
    \label{fig:udf-framework}
\end{figure}

\paragraph{Components.}
A policy \(\pi\) consists of five components (Figure~\ref{fig:udf-framework}):
\begin{enumerate}[leftmargin=*]
    \item \textbf{Local policy}: transforms logits through proposal, gating, scoring, or allocation, covering greedy decoding, temperature sampling, top-\(k\), top-\(p\), min-\(p\), and typical sampling.
    \item \textbf{Controller}: specifies the search topology, including autoregressive, beam-like, branching, tree-style, or resampling controllers.
    \item \textbf{Evaluator}: scores partial or complete candidates using likelihood, rollout utility, or an external verifier.
    \item \textbf{Transition}: updates the state by appending, expanding, branching, or resampling.
    \item \textbf{Scheduler}: allocates budget across samples, branches, or resampling steps.
\end{enumerate}

\paragraph{Evaluation metrics.}
We summarize each candidate set with four metrics: pass@\(k\) for support, SC for majority-vote concentration, best-of-\(N\) for evaluator-mediated selection, and FFS for early valid-answer success.
These metrics are not part of \udf{}; they provide complementary behavioral views of the same generated set.

\paragraph{Policy geometry.}
For structured analyses, we associate each operating point with
\begin{equation}
    z(u)=[P(u),Q(u),T(u),D(u),C(u)],
\end{equation}
where the coordinates denote pass@\(k\), SC, first-finish/validity, diversity or the pass--SC gap, and cost.
We combine component, behavioral, and budget distances:
\begin{equation}
\begin{aligned}
    d(u_i,u_j) = {} & \lambda_{comp}\, d_{comp}(\pi_i,\pi_j) \\
                    & + \lambda_z\|z(u_i)-z(u_j)\|_1 \\
                    & + \lambda_b|\log n_i-\log n_j|.
\end{aligned}
\end{equation}
This distance is used only for path diagnostics, not for mechanistic identification.

\paragraph{From SearchLens to SearchPath.}
SearchLens defines the operating space; SearchPath/BOPTR asks whether the RL curve can be recovered as a low-dimensional trajectory through it.
Appendix~\ref{app:c1} verifies that this five-component parameterization instantiates eleven mainstream decoding algorithms against vLLM, HuggingFace, and MCMC reference implementations.

\section{Experimental Setup}
\label{sec:setup}

We use Qwen2.5-7B/Math500 as the anchor cell for constructing the SearchLens landscape, recovery analysis, and BOPTR rule.
Other model--benchmark cells test horizontal transfer, vertical transfer, and cross-family stress settings, as summarized in Table~\ref{tab:blocks}.

\paragraph{Model pairs.}
We analyze paired Base/RL checkpoints from SimpleRL-Zoo, including Qwen2.5-0.5B\slash1.5B\slash7B\slash14B, Qwen2.5-Math-7B, Llama3.1-8B, and Mistral-7B-v0.1 when compatible generations are available.
Models trained on the SimpleRL-Zoo Hard RL-training split form the upper cohort; cross-family or weak-capacity models form the lower cohort.

\paragraph{Benchmarks.}
We evaluate Math500 \citep{hendrycks2021math, lightman2023verify}, AIME 2024/2025, GPQA-Diamond \citep{rein2023gpqa}, and IFEval \citep{zhou2023ifeval}.
IFEval is treated as an instruction-following and termination stress test, with pass rate, validity, and first-finish success as primary metrics and SC used diagnostically.

\paragraph{Policies, metrics, and cost.}
For each model and benchmark, we evaluate token-level autoregressive policies and, when available, expanded search policies such as MCMC-like resampling, entropy branching, and tree-style controllers, with budgets \(1,2,4,8,16\).
Our main analyses use same-policy curves, Base+\udf{} envelopes, near-match policies, and low-complexity rule families; tables report MAE in percentage points relative to the RL default-policy target.
The anchor cell requires approximately \(461\) H100 GPU-hours, dominated by eight expanded-controller runs.
Generations are deduplicated at \(b=16\) and reused across smaller budgets and all four metrics.

\begin{table}[t]
\centering
\small
\setlength{\tabcolsep}{4pt}
\begin{tabular}{llp{0.40\linewidth}}
\toprule
Block & Settings & Role \\
\midrule
Anchor & Qwen2.5-7B / Math500 & Full landscape construction and rule discovery. \\
Horizontal & Qwen2.5-7B / AIME, GPQA, IFEval & Same model, new benchmarks. \\
Vertical & Qwen2.5 family scales & Same family, different scale. \\
Stress & Llama3.1-8B, protocol-sensitive models & Cross-family and generation-protocol stress tests. \\
\bottomrule
\end{tabular}
\caption{Evaluation blocks. We report horizontal and vertical results separately because they test different transfer hypotheses.}
\label{tab:blocks}
\end{table}

\section{From Same-Policy Gains to Base Search Recovery}
\label{sec:anchor-findings}

We analyze the Qwen2.5-7B/Math500 anchor cell in two steps.
First, same-policy comparisons test whether RL changes behavior under a fixed decoding policy and budget.
Second, Base+\udf{} recovery tests whether the RL default-policy curve lies inside the base model's decoding/search landscape.
Together, these establish the prerequisites for SearchPath: a measurable RL shift and a Base+\udf{} support region that contains it.

\subsection{Same-Policy Gains: Where RL Improves Base}
\label{sec:samepolicy}

Same-policy comparison isolates post-training effects from gains due to richer inference-time policies.
Let \(A_M^{c}(\pi,b)\) denote the accuracy of model \(M\) under metric \(c\), policy \(\pi\), and budget \(b\).
We define
\begin{equation}
    \Delta_{same}^{c}(b)=A_{\rl}^{c}(\pi_0,b)-A_{\base}^{c}(\pi_0,b),
\end{equation}
where \(\pi_0\) is the RL default policy.

\begin{figure}[t]
    \centering
    \includegraphics[width=0.7\linewidth]{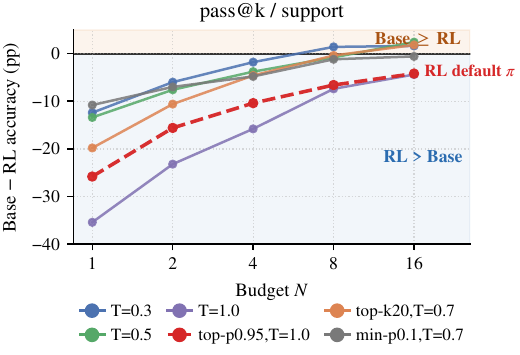}
    \caption{
        Same-policy Base--RL gap on Qwen2.5-7B/Math500 under pass@\(k\).
        Each line fixes a decoding policy \(\pi\) and plots
        \(A_{\base}(\pi,N)-A_{\rl}(\pi,N)\) over budgets \(N\in\{1,2,4,8,16\}\).
        Positive values indicate Base \(\geq\) RL.
        The dashed red line is the RL default policy.
        Base catches up only at larger budgets: the number of Base-shared policies with Base \(\geq\) RL is
        0/24, 0/24, 0/24, 2/24, and 13/24 across increasing budgets.
    }
    \label{fig:same-policy-lines}
\end{figure}

\paragraph{Observed patterns.}
Figure~\ref{fig:same-policy-lines} shows three patterns.
RL improves low-budget pass@\(k\) and SC on math-style tasks; Base can recover or exceed RL pass@\(k\) at larger budgets; and GPQA/IFEval shift metric semantics, with GPQA behaving more like concentration and IFEval better analyzed through validity and first-finish success.

\subsection{Base+Search Envelopes: Behavioral Recoverability}
\label{sec:recovery}

We next ask whether external decoding/search over the base model can approximate the RL default-policy curve.
This is a behavioral recoverability question, not a claim that RL implements an external policy.

\begin{figure}[t]
    \centering
    \includegraphics[width=0.7\linewidth]{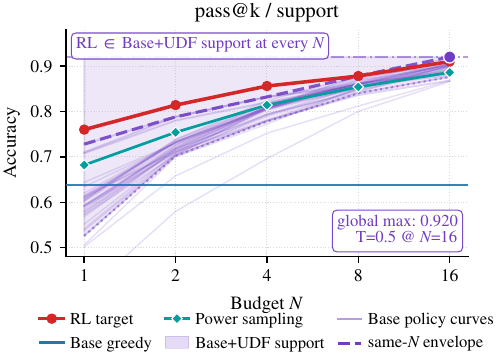}
    \caption{
        Per-budget Base+\udf{} support region on Qwen2.5-7B/Math500 under pass@\(k\).
        Thin curves show Base policy trajectories; the shaded region spans the Base policy cloud and the Base+\udf{} upper envelope.
        Red denotes the RL target, blue Base greedy, and teal the power-sampling baseline.
        The RL default target lies inside the Base+\udf{} support region at every budget.
    }
    \label{fig:support-region}
\end{figure}

\paragraph{Per-budget support region.}
For each metric \(c\) and budget \(b\), define the Base+\udf{} upper envelope:
\begin{equation}
    E_{\base}^{c}(b)=\max_{\pi\in\policyspace,\;n\le b} A_{\base}^{c}(\pi,n).
\end{equation}
Together with the low-quantile boundary of the Base policy cloud, this forms the support region in Figure~\ref{fig:support-region}.
The recovery gap,
\[
G^{c}(b)=A_{\rl}^{c}(\pi_0,b)-E_{\base}^{c}(b),
\]
is negative when Base+\udf{} exceeds the RL target and positive but small when the RL curve lies near the Base+\udf{} ceiling.
The RL target lies inside this support region at every budget.

\paragraph{Near-match sets and recovery paths.}
Because exact matches can arise post hoc in a large policy space, we define near-match sets
\begin{equation}
\begin{aligned}
    \mathcal{N}_{\epsilon}^{c}(b)=\{(\pi,n):\;& |A_{\base}^{c}(\pi,n) \\
    & - A_{\rl}^{c}(\pi_0,b)|\le \epsilon\},
\end{aligned}
\end{equation}
and select one closest Base+\udf{} candidate for each RL budget to form a recovery path.
This distinguishes structured recovery from unrelated per-budget lookup.
This distinction is important because recovery is not a same-budget comparison.
A Base model may underperform RL under the same default policy and budget, yet still contain Base+\udf{} operating points that match the RL target when policy and budget are allowed to vary.
Conversely, an envelope match alone is not sufficient evidence for a structured explanation.
We therefore treat near-match sets as the feasible region, and use the selected recovery path to test whether the matches form a coherent trajectory rather than isolated post-hoc choices.

\begin{figure}[!tb]
    \centering
    \includegraphics[width=0.78\textwidth]{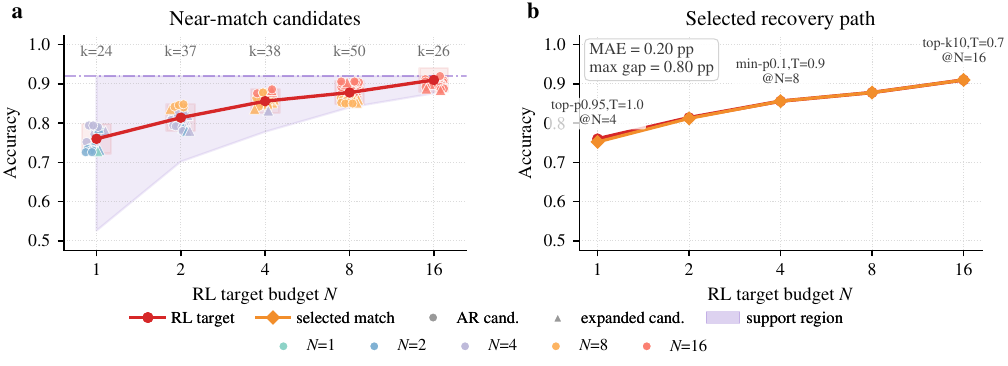}
    \caption{
        Equal-performance recovery on Qwen2.5-7B/Math500 under pass@\(k\).
        \textbf{(A)} Base+\udf{} near-match candidates within tolerance of each \(\mathrm{RL}@N_{\rl}\) target.
        \textbf{(B)} The selected recovery path closely tracks the RL curve, with residual gaps shown in gray.
    }
    \label{fig:recovery}
\end{figure}

Appendix Table~\ref{tab:recoveryobjects} summarizes each recovery object and its main risk.

\section{SearchPath: Low-Dimensional Recovery Paths}
\label{sec:boptr}

The recovery analysis in Section~\ref{sec:recovery} shows that RL targets have multiple Base+\udf{} matches at each budget.
SearchPath asks whether these matches form a transferable path rather than independent lookup results.
We formalize this path as a Budgeted Operating-Point Transition Rule (BOPTR).
BOPTR-P1, the pass@\(k\) instance, uses two structural assumptions: the budget exponent \(\beta_{g(d)}\) is tied to the benchmark regime \(g(d)\), while model-specific variation is captured by a scalar offset \(\delta_m\).

\subsection{From Matching Sets to Paths}
\label{sec:boptr-setup}

We define a \emph{cell} as a model--benchmark pair \((m,d)\), and let \(\mathcal{B}=\{1,2,4,8,16\}\) be the allowed budget set.
Let \(\operatorname{round}_{\mathcal{B}}\) denote rounding to the nearest allowed budget.
For each cell and channel \(\mathrm{ch}\in\{P,Q\}\), SearchPath predicts the RL pass rate through a Base operating point:
\begin{align}
\widehat{N}_{m,d}^{\mathrm{ch}}(b)
&=
\operatorname{round}_{\mathcal{B}}\!
\left(
\alpha_{m,d}^{\mathrm{ch}}\,
b^{\beta_{g(d)}^{\mathrm{ch}}}
\right),
\label{eq:boptr-N}
\\
\widehat{\pi}_{m,d}^{\mathrm{ch}}
&=
\argmin_{\pi}\,
\mathcal{L}_{m,d}^{\mathrm{ch}}(\pi),
\label{eq:boptr-pi}
\\
\widehat{P}_{\rl}^{m,d}(b)
&=
P_{\base}^{m,d}
\!\left(
\widehat{\pi},
\widehat{N}(b)
\right).
\label{eq:boptr-P}
\end{align}

The selector loss is
\begin{equation}
\begin{aligned}
\mathcal{L}_{m,d}^{\mathrm{ch}}(\pi)
&=
\sum_{b\in\mathcal{B}}
\left\|
r(\pi,\widehat{N}(b))
-
\bar{r}_{\mathrm{proto}}
\right\|_{W}
\\
&\quad
+
\lambda_{\mathrm{cost}}\,C(\pi).
\end{aligned}
\label{eq:boptr-loss}
\end{equation}
Here \(r=[\mathrm{rank}_P,\allowbreak\mathrm{rank}_{P-Q},\allowbreak
\mathrm{rank}_Q,\allowbreak\mathrm{ctrl},\allowbreak\log\mathrm{cost}]\), where \(\mathrm{ctrl}\) is the controller index.
We use \(W=(1,1,1,0.5,0.2)\) and \(\lambda_{\mathrm{cost}}=0.01\).
The scale parameter is decomposed as
\(\log\alpha_{m,d}=\mu_{g(d)}+\delta_m\), with \(\delta_m\) a model-specific scalar.
Different experimental protocols instantiate \(\delta_m\) differently; Section~\ref{sec:boptr-vertical} uses this decomposition to separate direct transfer, one-cell calibration baselines, and oracle fits.

We measure rule quality by
\(\mathrm{mech}_P\equiv|\widehat{P}_{\rl}-P_{\rl}|\cdot 100\) in percentage points.
Because each RL target \(P_{\rl}(b)\) already has multiple Base+\udf{} matches within \(\pm 3\) pp (Appendix~\ref{app:boptr-supp}), the rule's role is path selection rather than match discovery.

\subsection{Cross-Benchmark Transfer: An Empirical Budget Transition Rule}
\label{sec:boptr-horizontal}

We first test whether \(\beta\) is better explained by the model \(m\) or by the benchmark regime \(g(d)\).
Table~\ref{tab:regime-horizontal}A fits \(\beta\) on the anchor model under three policy constraints: R0 allows \(\pi\) to vary across budgets, R1 locks one \(\pi\) across all budgets, and R2 uses the default policy.
Agreement across these settings indicates that the budget relation is not merely an artifact of per-budget policy switching.

This analysis yields three regimes on the benchmarks we test.
Math500 is sublinear and policy-sensitive, so we set \(\beta_{\text{math}}=0.60\).
GPQA and IFEval are approximately linear, giving the shared nonmath/OOD regime \(\beta_{\text{nonmath}}=1.00\).
AIME fits \(\beta=0\) when the policy is locked, and we label this the floor regime with \(\beta_{\text{floor}}=0.00\).
The label is descriptive rather than a capability claim: refitting AIME into the math regime (\(\beta=0.60\)) changes the AIME cell error by \(+1.67\)\,pp with a 95\% CI of \([-5.00, +4.00]\) that contains zero, below the \(2.22\)\,pp AIME seed standard deviation, so the two maps are statistically indistinguishable on this grid.
We keep the pre-registered three-regime map for every headline number and report the two-regime fit only as a diagnostic counterfactual.
The taxonomy is likewise descriptive for the recipe and cohort we test: the number of regimes (3) is determined by the four available benchmarks and the minimal AIME sample size, and future additions of benchmarks may necessitate a re-categorization of regimes.

Table~\ref{tab:regime-horizontal}B evaluates this regime map against horizontal baselines.
BOPTR-P1 with regime-specific \(\beta\) is the strongest non-oracle rule, reducing the mean error from 6.01 pp for anchor per-budget copy to 3.41 pp.
The lower error is associated with rescaling \(N\) differently for OOD-nonmath cells, where directly copying the Math500 scaling would under-allocate the sample budget.
The remaining gap to the oracle smooth path reflects the cost of predicting \((\alpha,\pi)\) rather than fitting them separately for each cell.

\paragraph{Uncertainty and replication.}
A question-level paired bootstrap puts a 95\% CI of \([2.32, 5.53]\) on the 3.41\,pp anchor mean.
The paired contrasts against H2, H3, and H4 exclude zero (\(-4.69\)\,pp, \([-6.18, -2.75]\); \(-2.59\)\,pp, \([-4.09, -1.26]\); \(-3.15\)\,pp, \([-5.00, -1.80]\)), whereas H1 is at parity (\(-0.04\)\,pp, \([-2.69, +2.50]\)).
Repeating the pipeline under three decoding seeds gives \(3.07 \pm 0.39\)\,pp for the full rule and \(3.09 \pm 0.40\)\,pp when the selected policy is held fixed.
The policy selector operates on per-benchmark AR pools (23--24 policies for Math500, GPQA, and IFEval; 6 for AIME, whose remaining submission entries are structured decoders outside this runner), and all three seeds are scored on a common pool, so the reported spread reflects decoding rather than pool composition (Appendix Table~\ref{tab:reb-horizontal}).
Extending the budget grid to \(b=64\), and to \(b=256\) on AIME, raises the mean error to 5.02\,pp, concentrated on AIME at 12.50\,pp; under the frozen floor map the rule keeps \(\widehat{N}\equiv 2\), so the large-budget misfit is an allocation error of the budget map rather than a ceiling of base search.
Appendix Table~\ref{tab:reb-horizontal} reports these rows in full; recovery numbers from the AR-only and expanded-controller pools are never compared against each other (see Limitations).

\begin{table}[!t]
\scriptsize
\setlength{\tabcolsep}{2pt}
\renewcommand{\arraystretch}{0.96}

\begin{tabular}{@{}p{\twopanel}@{\hspace{\columnsep}}p{\twopanel}@{}}
\begin{minipage}[t]{\linewidth}
\vspace{0pt}
\raggedright
\begin{adjustbox}{max width=\linewidth}
\begin{tabular}{@{}lcccc@{}}
\toprule
benchmark & regime & R0 \(\beta\) & R1 \(\beta\) & R2 \(\beta\) \\
\midrule
Math500 & math & 0.50 & 0.70 & 0.50 \\
AIME & floor & 0.60 & 0.00 & 0.00 \\
GPQA & nonmath & 0.80 & 1.00 & 0.70 \\
IFEval & nonmath & 0.80 & 1.00 & 0.70 \\
\bottomrule
\end{tabular}
\end{adjustbox}

\panelnote{\textbf{(A)} Regime discovery. R0 allows \(\pi\) to vary across budgets, R1 locks one \(\pi\), and R2 uses the default \(\pi\).}
\end{minipage}
&
\begin{minipage}[t]{\linewidth}
\vspace{0pt}
\raggedright
\begin{adjustbox}{max width=\linewidth}
\begin{tabular}{@{}lccccc@{}}
\toprule
baseline & Math500 & AIME & GPQA & IFEval & mean \\
\midrule
H1 base greedy & 5.32 & --- & 4.24 & 1.55 & 3.70 \\
H2 base same-\(\pi\) & 12.52 & 5.00 & 8.69 & 6.21 & 8.10 \\
H3 anchor per-\(b\) copy & 0.20 & 1.67 & 15.46 & 6.70 & 6.01 \\
H4 strict BOPTR (\(\beta\!=\!0.6\)) & 1.12 & 1.00 & 17.17 & 6.95 & 6.56 \\
\textbf{H5 BOPTR-P1 (regime \(\beta\))} & \textbf{1.12} & \textbf{2.67} & \textbf{6.77} & \textbf{3.11} & \textbf{3.41} \\
H6 2D oracle per-\(b\) & 0.20 & 0.00 & 0.20 & 0.33 & 0.19 \\
H7 2D oracle smooth (R1 path) & 1.12 & 1.00 & 1.92 & 1.55 & 1.40 \\
\bottomrule
\end{tabular}
\end{adjustbox}

\panelnote{\textbf{(B)} Horizontal transfer. H5 is the strongest non-oracle rule.}
\end{minipage}
\end{tabular}

\vspace{0.4ex}

\caption{\footnotesize
Regime discovery and horizontal transfer on Qwen2.5-7B.
Errors are mean \(|\widehat{P}_{\rl}-P_{\rl}|\) in percentage points over \(b\in\{1,2,4,8,16\}\).
BOPTR-P1 is the strongest non-oracle horizontal rule, outperforming both direct anchor copying and strict Math500 scaling.
}
\label{tab:regime-horizontal}

\vspace{0.8ex}

\begin{tabular}{@{}p{\twopanel}@{\hspace{\columnsep}}p{\twopanel}@{}}
\begin{minipage}[t]{\linewidth}
\vspace{0pt}
\raggedright
\begin{adjustbox}{max width=\linewidth}
\begin{tabular}{@{}lccccc@{}}
\toprule
\multicolumn{6}{@{}l}{\textit{Upper: same Hard RL-training split}} \\
baseline & 7B & 14B & 1.5B & Math-7B & mean \\
\midrule
H1 base greedy & 3.71 & 4.16 & 15.47 & 6.63 & 7.49 \\
H2 base same-\(\pi\) & 8.10 & 8.29 & 7.07 & 7.18 & 7.66 \\
H3 anchor per-b copy & 6.01 & 6.92 & 14.24 & 7.58 & 8.69 \\
V1 per-cell oracle & 1.40 & 2.26 & 7.94 & 2.09 & 3.42 \\
V2 anchor path (no target-RL info) & 1.40 & 2.55 & 10.96 & 3.73 & 4.66 \\
\bottomrule
\end{tabular}
\end{adjustbox}

\vspace{0.45ex}

\begin{adjustbox}{max width=\linewidth}
\begin{tabular}{@{}lcccc@{}}
\toprule
\multicolumn{5}{@{}l}{\textit{Lower: cross-family / weak-capacity}} \\
baseline & 0.5B & Llama & Mistral & mean \\
\midrule
H1 base greedy & 3.93 & 9.06 & 5.79 & 6.26 \\
H2 base same-\(\pi\) & 7.73 & 8.37 & 8.09 & 8.06 \\
H3 anchor per-b copy & 8.32 & 9.79 & 9.44 & 9.18 \\
V1 per-cell oracle & 1.41 & 2.56 & 1.70 & 1.89 \\
V2 anchor path (no target-RL info) & 4.92 & 10.16 & 3.83 & 6.30 \\
\bottomrule
\end{tabular}
\end{adjustbox}

\panelnote{\textbf{(A)} Direct anchor-path vertical baselines without target-RL information. Each cell reports mean mech\(_P\) in percentage points over budgets.}
\end{minipage}
&
\begin{minipage}[t]{\linewidth}
\vspace{0pt}
\raggedright
\begin{adjustbox}{max width=\linewidth}
\begin{tabular}{@{}>{\raggedright\arraybackslash}p{0.34\linewidth}
                >{\raggedright\arraybackslash}p{0.42\linewidth}
                >{\centering\arraybackslash}p{0.18\linewidth}@{}}
\toprule
rule & \makecell{structural\\assumption} & \makecell{overall\\mean (pp)} \\
\midrule
V1 per-cell oracle & none (fit \(\alpha,\beta,\pi\) per cell) & 2.77 \\
V2 \(\delta_m\!=\!0\) & formula shared, no offset & 5.36 \\
V3 \(\delta_m\) from Math500 & formula + 1-D offset & 4.44 \\
\textbf{V5b base-only \(\hat{\delta}_m\)} & formula + predicted 1-D offset & \textbf{4.19} \\
\midrule
\multicolumn{3}{@{}l}{\textit{RL-information frontier}} \\
Z1-\(N^{\star}\) one RL anchor cell & \(\alpha\) from anchor saturation knee & 4.04 \\
Z0-\(u\) no RL at all & \(\alpha\) from base@rltrain fingerprint & 5.08 \\
\bottomrule
\end{tabular}
\end{adjustbox}

\panelnote{\textbf{(B)} Formula decomposition and RL-information frontier. V2\(\to\)V3 measures the value of \(\delta_m\) calibration, V3\(\to\)V5b replaces that calibration with a base-only prediction, and V3\(\to\)V1 measures the residual structural cost. The lower block reduces RL access further, to one anchor cell and then to none. Macro means over seven models; Appendix Table~\ref{tab:reb-vertical} adds the micro means over cells.}
\end{minipage}
\end{tabular}

\vspace{0.4ex}

\caption{\footnotesize
Vertical transfer and formula decomposition across seven models.
Panel A tests direct anchor-path transfer (V2) without target-RL information.
Panel B decomposes the formula: V1 per-cell oracle, V2 with no model offset, V3 with a one-dimensional offset \(\delta_m\) estimated from each model's Math500 Base/RL pair, and V5b with that offset predicted from base-only statistics (fit on the other cohort models by leave-one-model-out).
The lower block of Panel B traces the RL-information frontier as target-RL access shrinks: Z1-\(N^{\star}\) uses a single RL anchor cell and Z0-\(u\) uses no RL at all, reading \(\alpha\) from base rollouts on the RL-training distribution.
Means are macro (over seven per-model means); Appendix Table~\ref{tab:reb-vertical} co-reports micro means over cells.
}
\label{tab:vertical-combined}
\end{table}

\subsection{Cross-Model Transfer: Shared Form, Model-Specific Realization}
\label{sec:boptr-vertical}

We next test vertical transfer without target-RL information.
Given the anchor \((\pi,N)\) path, Table~\ref{tab:vertical-combined}A asks how well the rule predicts \(P_{\rl}\) on new models.
Direct anchor-path transfer works within the same Hard RL-training split at matched or comparable scale: the 7B, 14B, and Math-7B models obtain errors of 1.40, 2.55, and 3.73 pp.
It degrades for the 1.5B downscale and for the cross-family Llama setting, where the error reaches 10.16 pp.
This suggests that the shared RL-training distribution is important for direct \((\pi,N)\) transfer, even when the formula family remains useful.

\paragraph{V3 as a low-dimensional offset baseline.}
V3 estimates one scalar \(\delta_m\) from each model's Math500 Base/RL pair and then reuses the same regime exponent and policy selector on the remaining benchmarks.
As shown in Table~\ref{tab:vertical-combined}B, this one-dimensional offset reduces the overall error from 5.36 to 4.44 pp.
The improvement is especially pronounced in the lower cohort, where the mean error drops from 6.30 to 4.21 pp and the Llama error drops from 10.16 to 3.68 pp.
This supports the decomposition \(\log\alpha_{m,d}=\mu_{g(d)}+\delta_m\) as an informative structural form.

The remaining gap from V3 to the per-cell oracle is 1.7 pp.
This residual reflects the cost of using one \(\beta\) per regime and a behavior-coordinate selector instead of fitting each cell directly.
Across the seven models, \(\alpha_{\text{math}}\) spans a factor of 3.0 and the R1-locked \(\pi_{\text{math}}\) varies by model (Appendix Table~\ref{tab:formula-same}), but much of this variation is absorbed by \(\delta_m\).
A base-only predicted offset, fit on the other cohort models by leave-one-model-out and using no Math500 RL calibration for the target model, reaches 4.19\,pp (V5b in Table~\ref{tab:vertical-combined}B), on par with the RL-calibrated V3, with its improvement over V3 concentrated on the 1.5B model.

\paragraph{Coverage beyond the fitted cohort.}
We apply the frozen rule to three RL checkpoints trained outside the SimpleRL-Zoo recipe, which takes the cohort from seven to ten models across four families.
Open-Reasoner-Zero-7B, OAT-7B, and DeepSeek-Math-7B give 4.87, 4.48, and 3.28 pp, inside the range of the fitted cohort (Table~\ref{tab:coverage-beyond}A).
We also freeze the regime exponents, the policy selector, and the submission-era \(\delta_m\), and evaluate on four benchmarks the rule was never fitted on: AMC23, Minerva, OlympiadBench, and TinyMMLU.
Across the same seven models this held-out grid gives 5.03 pp against 4.44 pp in fit, and all four benchmark means stay under the pre-registered 8 pp criterion, although five individual cells exceed it (Table~\ref{tab:coverage-beyond}B).
Appendix Table~\ref{tab:reb-vertical} adds per-model V3 and V5b rows and the construction details for the added checkpoints.

\begin{table}[!t]
\scriptsize
\setlength{\tabcolsep}{3pt}

\begin{tabular}{@{}p{\twopanel}@{\hspace{\columnsep}}p{\twopanel}@{}}
\begin{minipage}[t]{\linewidth}
\vspace{0pt}
\raggedright
\begin{adjustbox}{max width=\linewidth}
\begin{tabular}{@{}lccc@{}}
\toprule
\multicolumn{4}{@{}l}{\textit{New models: second RL recipes and a cross-family model}} \\
rule & ORZ-7B & OAT-7B & DS-Math-7B \\
\midrule
V3 \(\delta_m\) from Math500 & 4.87 & 4.48 & 3.28 \\
V5b base-only \(\hat{\delta}_m\) & 4.87 & 4.21 & --- \\
V3 pipeline, \(\delta_m\!:=\!0\) & 4.68 & 4.21 & 3.28 \\
\bottomrule
\end{tabular}
\end{adjustbox}

\panelnote{\textbf{(A)} The frozen rule on three RL checkpoints trained outside
the SimpleRL-Zoo recipe: Open-Reasoner-Zero-7B, OAT-7B, and the cross-family
DeepSeek-Math-7B.
V5b is undefined for DS-Math-7B, which lies outside the leave-one-model-out
cohort; the \(\delta_m\!:=\!0\) row is the zero-calibration reference.
Construction details in Appendix Table~\ref{tab:reb-vertical}.}
\end{minipage}
&
\begin{minipage}[t]{\linewidth}
\vspace{0pt}
\raggedright
\begin{adjustbox}{max width=\linewidth}
\begin{tabular}{@{}lccccc@{}}
\toprule
\multicolumn{6}{@{}l}{\textit{Held-out benchmarks \(\times\) models, frozen rule}} \\
model & AMC23 & Minerva & Olymp. & TinyMMLU & mean \\
\midrule
7B (anchor) & 3.00 & 3.75 & 2.84 & 1.60 & 2.80 \\
14B & 5.00 & 3.09 & 2.79 & 4.60 & 3.87 \\
1.5B & 13.50 & 6.25 & 6.67 & 11.20 & 9.40 \\
Math-7B & 4.00 & 8.16 & 2.70 & 7.00 & 5.46 \\
0.5B & 6.00 & 1.69 & 2.93 & 2.40 & 3.25 \\
Llama & 10.00 & 4.34 & 2.58 & 6.60 & 5.88 \\
Mistral & 5.00 & 3.31 & 1.57 & 8.40 & 4.57 \\
\midrule
benchmark mean & 6.64 & 4.37 & 3.15 & 5.97 & 5.03 \\
\bottomrule
\end{tabular}
\end{adjustbox}

\panelnote{\textbf{(B)} V3-frozen transfer to four held-out benchmarks
(\(\mu_g\), \(\beta_g\), selector, and submission-era \(\delta_m\) frozen;
regimes assigned a priori, Appendix Table~\ref{tab:reb-horizontal}A;
\(n=40/272/675/{\approx}100\) questions; all 28 cells complete).
Bottom-right corner = mean over the 28 cells (macro and micro coincide on the
balanced grid); per-cell oracle on the same cells 3.69, frozen-minus-oracle gap
1.34\,pp.
All four benchmark means meet the pre-registered \(\le\)8\,pp criterion; five
individual cells exceed it (1.5B--AMC23 13.50, 1.5B--TinyMMLU 11.20,
Llama--AMC23 10.00, Mistral--TinyMMLU 8.40, Math-7B--Minerva 8.16), all
weak-capacity, cross-family, or math-specialized cells.}
\end{minipage}
\end{tabular}

\vspace{0.4ex}

\caption{\footnotesize
Coverage beyond the fitted cohort, with all frozen parameters and no refitting.
(A) Three RL checkpoints from recipes other than SimpleRL-Zoo, taking the cohort
to ten models across four families.
(B) The frozen rule on four benchmarks it was never fitted on, across the seven
cohort models.
The Math-7B RL cells, previously excluded for degenerate generations, were traced
to a corrupted local weight copy and regenerated from healthy weights.
A fifth held-out benchmark, MMLU-STEM (\(n{=}3153\)), has base-side generations
only (no RL counterpart in this cohort) and enters the descriptive analysis only.
}
\label{tab:coverage-beyond}
\end{table}

\subsection{Takeaways}
\label{sec:boptr-takeaways}

BOPTR-P1 supports a hierarchical view of recovery.
The formula family is shared, with \(\beta\) determined by benchmark regime, but its realization is model-specific through \(\alpha\) and the selected policy.
Direct anchor-path transfer works best within the same SimpleRL-Zoo difficulty split and degrades across model families or training distributions.
A one-dimensional offset absorbs much of this variation, reducing the overall error to 4.44 pp, and a base-only predicted offset matches this without target-model RL calibration (4.19 pp).
Richer rule families that add free axes without matching training signal worsen transfer (Appendix~\ref{app:rule-variants}); a single base-only feature is too weak to predict \(\delta_m\) in the six-model cohort (Appendix~\ref{app:boptr-v4}), but a multi-signal base-only predictor recovers most of the offset (V5b, Table~\ref{tab:vertical-combined}B).
We therefore use BOPTR as a behavioral diagnostic for how RL relates to Base+\udf{} operating points under the SimpleRL-Zoo recipe, not as a universal post-RL performance predictor.

\section{Discussion and Conclusion}
\label{sec:discussion}

Our results give a behavioral interpretation of the phrase ``RL internalizes search.''
This does not mean that RL corresponds to a single named decoding policy, nor that the RL model literally implements an external search procedure.
Rather, under the tested SimpleRL-Zoo recipe, many RL default-policy curves can be approximated by structured Base+\udf{} operating paths.
BOPTR-P1 makes this relationship operational by selecting a Base tuple \((\widehat{\pi},\widehat{N})\), where \(\widehat{N}\) follows a regime-conditioned exponent and \(\widehat{\pi}\) is chosen by behavior-coordinate matching.

This view clarifies what recovery can and cannot explain.
A same-policy comparison may underestimate the relationship between Base and RL, because restricting evaluation to a single \((\pi_0,b)\) tuple hides nearby Base+\udf{} operating points that already match the RL target.
When a smooth Base+\udf{} path tracks the RL curve, the corresponding gain can be read as improved access to behavior already present in the base operating landscape.
When recovery fails, or when SC exposes a ceiling, the gap points to behavior not captured by the tested policy pool or by the selected behavioral channel.
Thus, \udf{} is useful not only for matching RL behavior, but also for localizing where such matching breaks.

The resulting picture is hierarchical.
The BOPTR formula family is shared across regimes, but its realization depends on both benchmark and model: task regime controls the budget exponent, while model-specific offsets and selected policies determine the concrete path.
This also suggests how the analysis could extend beyond SimpleRL-Zoo.
For another post-training recipe, the same operating-point view can ask which gains correspond to movement inside the base landscape and which require behavior outside the tested policy pool.
However, regime exponents and model offsets should be re-estimated rather than assumed to transfer unchanged.

\paragraph{Conclusion.}
We introduced SearchLens/\udf{} as a unified behavioral map for decoding and search, and SearchPath/BOPTR as a low-dimensional rule for tracing RL behavior through this map.
Our findings support a qualified internalized-search view: RL often improves access to behaviors that are already reachable from the base model under suitable inference-time policies, but the concrete recovery path is benchmark- and model-specific.
Cross-family degradation delimits this interpretation, and predicting \(\delta_m\) still relies on a cohort of RL models rather than on the target model alone: BOPTR is best used as a diagnostic of RL--Base relationships, not as a universal post-RL performance predictor.

\FloatBarrier

\section*{Acknowledgments}
This project was supported by the National Natural Science Foundation of China (NSFC) under Grant No.~62576102.

\section*{Limitations}
\label{sec:limitations}

\paragraph{Behavioral, not parameter-level, equivalence.}
Our recoverability and rule results are behavioral statements about the operating-point landscape: they show that Base+\udf{} contains operating points whose pass-rates and concentration match RL targets, and that a small low-degree rule selects such points.  This is not a claim that the RL model literally executes any external decoding policy; we make no parameter-level mechanistic identification.

\paragraph{Policy-pool comparability.}
Some settings expose AR-only pools while others include expanded controllers (MCMC-like resampling, entropy branching, tree controllers).  Recoverability statements should be interpreted relative to the \udf{} policies actually evaluated in each cell; we mark AR-only and expanded-pool cells separately and never compare envelopes across mismatched pools.

\paragraph{Metric heterogeneity across benchmarks.}
Math500, AIME, GPQA, and IFEval do not all admit the same evaluation-metric semantics.  IFEval is treated as an evaluation-metric mismatch / OOD stress test, with first-finish and validity as primary metrics and SC as diagnostic; GPQA often behaves as a sharpening/concentration task rather than a support-expansion task.  Cross-benchmark numbers should be read with this regime structure in mind, not as a uniform metric.

\paragraph{Model-specific parameters.}
The BOPTR formula family is shared, but its parameters and selected policies are model-specific.  In our analysis we estimate the per-model offset \(\delta_m\) from each model's Math500 Base/RL pair (V3) as a one-cell calibration baseline that bounds how much of the cross-model gap is absorbable by a low-dimensional offset.  A base-only predictor that uses no target-model RL calibration (V5b) recovers most of this offset when fit leave-one-model-out on the cohort; a single-feature linear ridge does not, since at $n{=}6$ models it stays inside the $y$-shuffle null (Appendix~\ref{app:boptr-v4}).  The predictor still depends on a cohort of RL models rather than a single target, and its behavior beyond the tested cohort is not established.  We position the rule as a behavioral diagnostic under the tested SimpleRL-Zoo recipe, not as a universal post-RL performance predictor; stronger absolute-prediction or checkpoint-selection claims would require larger model cohorts, training-distribution calibration sets, or model-conditioned adapters beyond the current rule family.

\paragraph{Scope of model coverage.}
Results cover one Qwen-family scale sweep (0.5B/1.5B/7B/14B), one same Hard RL-training split cross-family check (Qwen2.5-Math-7B), one cross-family check (Llama-3.1-8B), and one weak-capacity check (Mistral-7B-v0.1) on paired Base/RL checkpoints from a single RL recipe (SimpleRL-Zoo).  We do not claim our regime map ($\beta_{\text{math}}{=}0.6$, $\beta_{\text{nonmath}}{=}1.0$, $\beta_{\text{floor}}{=}0$) transfers verbatim to other RL recipes or to substantially different model families without re-anchoring.  Training-duration effects at the scale of prolonged RL training \citep{liu2025prorl} remain untested; our claims are confined to the recipes and training durations evaluated.  The number of regimes (3) is likewise determined by the four available benchmarks and the minimal AIME sample size, and adding benchmarks may necessitate a re-categorization of regimes.

\FloatBarrier

\clearpage
\FloatBarrier
\clearpage
\appendix
\section{Appendix}
\suppressfloats[t]

\subsection{Notation}
\label{app:notation}

Table~\ref{tab:notation} lists every recurring symbol, one line each.

\begin{table}[H]
\centering
\footnotesize
\setlength{\tabcolsep}{3pt}
\renewcommand{\arraystretch}{1.05}
\begin{tabular}{@{}lL{0.66\columnwidth}@{}}
\toprule
symbol & meaning \\
\midrule
\(u=(\pi,n)\) & \udf{} operating point: policy \(\pi\) and sample budget \(n\) \\
\(\policyspace\) & \udf{} policy space (local policy, controller, evaluator, transition, scheduler) \\
\(\budgetset\) & allowed budget set \(\{1,2,4,8,16\}\) \\
\(b\) & RL-side rollout budget being predicted \\
\((m,d)\) & cell: model \(m\) on benchmark \(d\) \\
\(z(u)\) & behavior signature \([P(u),Q(u),T(u),D(u),C(u)]\): pass@\(k\), SC, first-finish/validity, diversity, cost \\
\(d(u_i,u_j)\) & operating-point distance \(\lambda_{comp}\,d_{comp}+\lambda_z\|\Delta z\|_1+\lambda_b|\Delta\log n|\) \\
\(g(d)\) & regime label of benchmark \(d\) (math / nonmath / floor) \\
\(\beta_{g(d)}\) & regime budget exponent (0.60 / 1.00 / 0.00) \\
\(\alpha_{m,d}\) & per-cell scale, \(\log\alpha_{m,d}=\mu_{g(d)}+\delta_m\) \\
\(\mu_{g}\) & regime-level component of \(\log\alpha\) \\
\(\delta_m\) & model-specific scalar offset \\
\(\operatorname{round}_{\budgetset}\) & rounding to the nearest allowed budget \\
\(\widehat{N}(b)\) & predicted base budget \(\operatorname{round}_{\budgetset}(\alpha b^{\beta})\) (Eq.~\ref{eq:boptr-N}) \\
\(\widehat{\pi}\) & selected policy, \(\argmin_\pi \mathcal{L}\) (Eq.~\ref{eq:boptr-pi}) \\
\(r,\ \bar{r}_{\mathrm{proto}}\) & selector features \([\mathrm{rank}_P,\) \(\mathrm{rank}_{P-Q},\) \(\mathrm{rank}_Q,\) \(\mathrm{ctrl},\) \(\log\mathrm{cost}]\) and anchor prototype \\
\(W,\ \lambda_{\mathrm{cost}}\) & selector weights \((1,1,1,0.5,0.2)\) and cost penalty \(0.01\) \\
\(\mathrm{ch}\) & prediction channel, \(\mathrm{ch}\in\{P,Q\}\) \\
\(\widehat{P}_{\rl}(b)\) & rule prediction \(P_{\base}(\widehat{\pi},\widehat{N}(b))\) (Eq.~\ref{eq:boptr-P}) \\
\(\mathrm{mech}_P\) & transfer error \(|\widehat{P}_{\rl}-P_{\rl}|\cdot 100\) (pp) \\
H1--H7 & horizontal baselines and oracles (Table~\ref{tab:regime-horizontal}B) \\
V1--V5b & vertical-transfer tiers (Table~\ref{tab:vertical-combined}B) \\
Z1-\(N^{\star}\), Z0-\(u\) & RL-information frontier rungs: one RL anchor cell / no RL at all \\
\bottomrule
\end{tabular}
\caption{\footnotesize Notation used throughout the paper, one line per symbol.}
\label{tab:notation}
\end{table}

\subsection{UDF expressiveness: equivalence to mainstream decoding algorithms}
\label{app:c1}

To verify that \udf{} is not just a notational wrapper, we instantiate eleven mainstream decoding algorithms as \udf{} operating points and check that they reproduce reference implementations on the same prompts (Math500 subset, $n{=}50$, Qwen2.5-7B, identical seed).  Tier 1--2 covers the vLLM \texttt{SamplingParams} family (greedy, temperature, top-$k$, top-$p$, min-$p$, and two repetition/frequency-penalty variants); Tier 3--4 covers external reference implementations (HuggingFace \texttt{typical\_p}, HuggingFace \texttt{num\_beams}, the official MCMC \texttt{power\_sampling}, and an internal \texttt{entropy\_tree} branching reference).  All Tier 1--2 strategies match their references with output match rate $\ge 0.76$ and EM agreement $\ge 0.92$ and are flagged equivalent under the pre-registered threshold (Table~\ref{tab:c1-tier12}).  Tier 3--4 strategies match within $|\Delta\mathrm{EM}|\le 0.08$ (Table~\ref{tab:c1-tier34}).  This establishes that the five-component \udf{} parameterization is empirically expressive enough to recover the decoding policies used in published baselines, so that downstream same-policy and recovery analyses operate on an instantiation-faithful policy space.

\begin{table}[H]
\centering
\small
\setlength{\tabcolsep}{4pt}
\begin{tabular}{lrrc}
\toprule
Strategy & Output match & EM agree & Equiv. \\
\midrule
greedy             & 0.86 & 0.92 & \checkmark \\
temperature        & 0.98 & 1.00 & \checkmark \\
top-$k$            & 0.92 & 0.96 & \checkmark \\
top-$p$            & 0.98 & 1.00 & \checkmark \\
min-$p$            & 0.92 & 1.00 & \checkmark \\
temp + rep.\ penalty   & 0.76 & 0.98 & \checkmark \\
top-$p$ + freq.\ penalty & 0.86 & 0.98 & \checkmark \\
\bottomrule
\end{tabular}
\caption{Tier 1--2 equivalence: \udf{} instantiation vs.\ vLLM \texttt{SamplingParams} reference on Math500 ($n{=}50$, Qwen2.5-7B).  Output match is the per-prompt string-match rate; EM agree is the agreement of extracted answers.  All seven strategies pass the pre-registered equivalence threshold.}
\label{tab:c1-tier12}
\end{table}

\begin{table}[H]
\centering
\small
\setlength{\tabcolsep}{3pt}
\begin{adjustbox}{max width=\linewidth}
\begin{tabular}{llrrr}
\toprule
Algorithm & Reference & EM ref & EM \udf{} & $|\Delta|$ \\
\midrule
typical          & HF \texttt{typical\_p}        & 0.36 & 0.44 & 0.08 \\
beam search      & HF \texttt{num\_beams}        & 0.52 & 0.50 & 0.02 \\
power sampling   & Official MCMC \texttt{power\_samp} & 0.50 & 0.48 & 0.02 \\
entropy branching & Internal \texttt{entropy\_tree} ref & 0.40 & 0.40 & 0.00 \\
\bottomrule
\end{tabular}
\end{adjustbox}
\caption{Tier 3--4 equivalence: \udf{} instantiation vs.\ external reference implementations on Math500 ($n{=}50$, Qwen2.5-7B).  $|\Delta|$ is $|\mathrm{EM}_{\rm ref}-\mathrm{EM}_{\udf{}}|$; for \texttt{power\_sampling}, the reference MCMC accept ratio is 0.657; for \texttt{entropy\_tree}, output match rate $=1.00$ (token-identical).  All four match within $|\Delta\mathrm{EM}|\le 0.08$.}
\label{tab:c1-tier34}
\end{table}

\subsection{Same-policy and Recovery supplements}
\label{app:anchor-supp}

\subsubsection{Qualitative same-policy signatures}
\label{app:signatures}

Table~\ref{tab:signatures} catalogues the four qualitative same-policy signatures used throughout the analysis.

\begin{table}[H]
\centering
\scriptsize
\setlength{\tabcolsep}{1.8pt}
\renewcommand{\arraystretch}{1.08}
\begin{tabularx}{\columnwidth}{@{}L{0.22\columnwidth}
                                L{0.18\columnwidth}
                                L{0.22\columnwidth}
                                Y@{}}
\toprule
Pattern & Typical metric & Example block & Interpretation \\
\midrule
Low-budget RL gain & pass@k, SC & Math500 / AIME & Sampling efficiency improves. \\
Base high-budget recovery & pass@k & GPQA / some math settings & Diversity/support remains in base. \\
Concentration gain & SC or SC / pass@k & GPQA & RL sharpens answer distribution. \\
Termination gain & FFS / validity & IFEval / long outputs & RL improves completion reliability. \\
\bottomrule
\end{tabularx}
\caption{Qualitative same-policy signatures used throughout the analysis.}
\label{tab:signatures}
\end{table}

\subsubsection{Recovery analysis objects}
\label{app:recoveryobjects}

Table~\ref{tab:recoveryobjects} lists the four recovery objects (envelope / near-match set / smooth path / frozen rule), what each one is allowed to show, and its main risk.

\begin{table}[!b]
\centering
\scriptsize
\setlength{\tabcolsep}{3pt}
\renewcommand{\arraystretch}{1.08}
\begin{tabularx}{0.98\textwidth}{@{}L{0.16\textwidth}
                                      L{0.20\textwidth}
                                      C{0.12\textwidth}
                                      L{0.22\textwidth}
                                      Y@{}}
\toprule
Analysis object &
Uses RL target? &
Allows lookup? &
Purpose &
Main risk \\
\midrule
Envelope &
no per-policy target &
yes &
Upper-bound support &
Overstates practical policy. \\
Near-match set &
yes &
yes &
Recoverability &
Too many matches. \\
Smooth path &
yes &
constrained &
Structured recovery &
Hyperparameter sensitivity. \\
Frozen rule &
train anchor only &
no &
Transfer diagnostic &
Under-identification. \\
\bottomrule
\end{tabularx}
\caption{Recovery analyses and what each one is allowed to show.}
\label{tab:recoveryobjects}
\end{table}

\subsubsection{Same-policy SC companion}
\label{app:same-policy-sc}

\begin{figure}[H]
    \centering
    \includegraphics[width=0.7\linewidth]{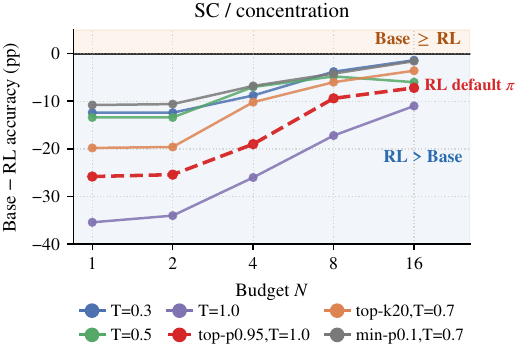}
    \caption{
        \textbf{Same-policy Base vs RL gap, Qwen2.5-7B / Math500 (SC).}
        Self-consistency (\texttt{majority\_vote@k}) companion of Figure~\ref{fig:same-policy-lines}.  Each line fixes one decoding policy~$\pi$ and plots $A_{\base}(\pi, N) - A_{\rl}(\pi, N)$ (pp).
    }
    \label{fig:same-policy-lines-sc}
\end{figure}

{\RaggedRight
Figure~\ref{fig:same-policy-lines-sc} is the self-consistency (\texttt{majority\_vote@k}) companion of main Figure~\ref{fig:same-policy-lines}.
\par}

\subsubsection{Support-region SC companion}
\label{app:support-region-sc}

\begin{figure}[H]
    \centering
    \includegraphics[width=0.7\linewidth]{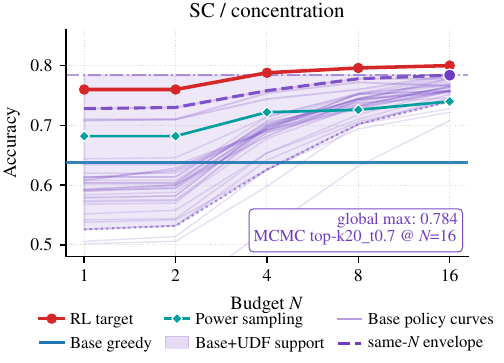}
    \caption{
        \textbf{Per-budget Base+\udf{} support region, Qwen2.5-7B / Math500 (SC).}
        Self-consistency (\texttt{majority\_vote@k}) companion of Figure~\ref{fig:support-region}.  Convention as in the main panel; the RL target remains inside the Base+\udf{} support region at every budget.
    }
    \label{fig:support-region-sc}
\end{figure}

{\RaggedRight
Figure~\ref{fig:support-region-sc} is the self-consistency (\texttt{majority\_vote@k}) companion of main Figure~\ref{fig:support-region}.
\par}

\subsubsection{Full same-policy gap map}
\label{app:same-policy-heatmap}

\begin{figure}[!t]
    \centering
    \includegraphics[width=0.7\linewidth]{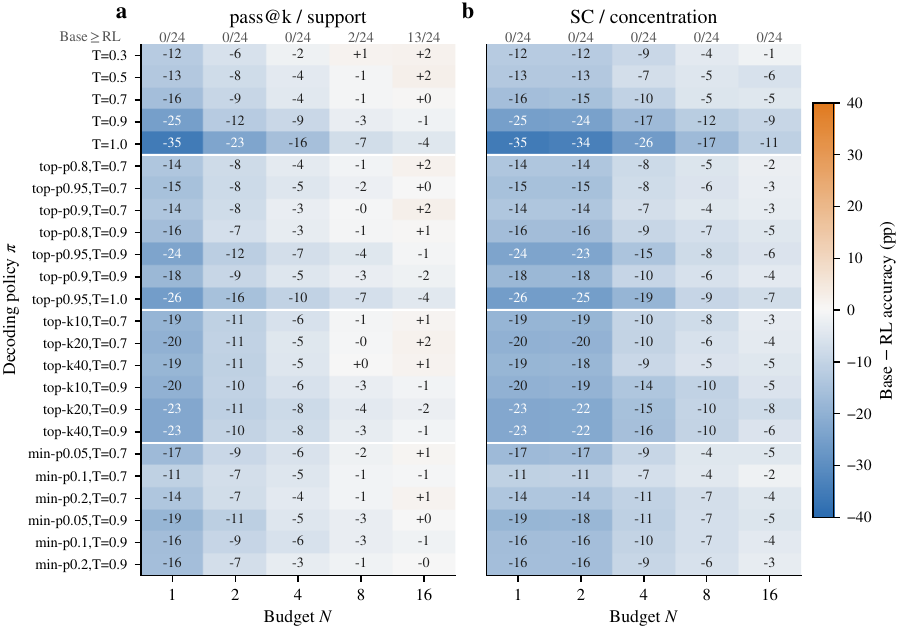}
    \caption{
        \textbf{Full same-policy gap map, Qwen2.5-7B / Math500.}
        Each cell shows $A_{\base}(\pi, N) - A_{\rl}(\pi, N)$ (percentage points)
        at the same policy $\pi$ and budget $N$.
        Rows are the 23 base-shared decoding policies, grouped by family
        (temperature, top-p, top-k, min-p; families separated by white horizontal lines
        and sorted within each family by the relevant hyperparameter).
        Columns are budgets $N \in \{1, 2, 4, 8, 16\}$.
        \textbf{(a)}~\texttt{pass@k}.
        \textbf{(b)}~self-consistency (\texttt{majority\_vote@k}).
        Blue cells: RL exceeds Base at the same $(\pi, N)$.
        Orange cells: Base~$\geq$~RL.
        The strip below each panel reports, at each budget, the number of policies
        with Base~$\geq$~RL.
        The diverging color scale is symmetric around zero; cell numbers are the gap
        rounded to the nearest percentage point.
        Companion to main Figure~\ref{fig:same-policy-lines}.
    }
    \label{fig:same-policy-heatmap}
\end{figure}

{\RaggedRight
Figure~\ref{fig:same-policy-heatmap} extends main Figure~\ref{fig:same-policy-lines} to every base-shared decoding policy (23 policies, grouped by family) at every budget.
\par}

\FloatBarrier

\subsubsection{UDF sweep compute cost on the anchor cell}
\label{app:udf-cost}

Table~\ref{tab:udf-cost} reports the wall-clock and token cost of producing the Qwen2.5-7B/Math500 UDF sweep used in Section~\ref{sec:anchor-findings}.  We separately tally three channels: (i)~\textbf{AR-Base} on the 23 base-shared policies plus the RL default \texttt{topp0.95\_t1.0} (24 policies total) under the base model; (ii)~\textbf{AR-RL} on the same 24 policies under the RL model; and (iii)~\textbf{Expanded base}, the cross-family controllers entered in the recovery near-match search (entropy\_branch with $b{=}4$ tree, sequence MCMC with $b{=}4$, MCTS with $16$ iterations, and thought tree with $b\in\{4,8\}$).  All runs use vLLM with tensor-parallel $=2$ on H100s; GPU-hours below are wall-hours $\times 2$.

\paragraph{Amortization across budgets and metrics.}
Each policy is generated once at the largest budget ($b{=}16$) and the same sample pool is then evaluated at every smaller budget $b\in\{1,2,4,8\}$ and at all four metrics (pass@$k$, best-of-$n$, majority-vote, first-finish).  We therefore deduplicate at the generation-run level (one row per AR policy; one row per (controller-mode, sampling-policy, internal-budget) for expanded controllers) and report the maximum wall-time within each group as the true compute cost.  The 5 budgets $\times$ 4 metrics per AR policy and the additional metric/budget combinations per expanded controller appear in the raw \texttt{strategy\_sweep.json} entries but do not add wall-clock cost beyond the single generation.

\paragraph{Headline numbers.}
The 24-policy AR sweep is cheap on both the base ($\approx 2.3$ wall-h, $\approx 124$M tokens) and the RL model ($\approx 2.1$ wall-h, $\approx 122$M tokens); roughly $4.6$ and $4.1$ H100 GPU-hours respectively under TP$=2$.  Expanded controllers dominate the budget: 8 deduped (controller, sampling-policy, internal-budget) runs consume $\approx 226$ wall-h ($\approx 452$ GPU-hours), $\approx 100\times$ the AR sweep, because each expanded sample is an internal search tree of multiple raw generations.  Across all three channels the anchor cell consumes $\approx 230$ wall-hours / $\approx 461$ H100 GPU-hours / $\approx 0.31$B tokens, which sets the per-cell cost we extrapolate from in the cross-model BOPTR study (Sections~\ref{sec:boptr-horizontal}--\ref{sec:boptr-vertical}).

\paragraph{What this excludes.}
The numbers above cover only the canonical post-paper sweeps (\texttt{sweep\_\{base,rl\}\_full} plus the two \texttt{f2\_*\_base\_full} expanded sweeps).  Exploratory smaller-question runs (\texttt{f0\_*}, \texttt{f1\_*} validation shards) and shard-level partial sweeps that were superseded by the canonical files are not included; they would roughly double the figure.  RL training cost itself is also excluded -- only the post-training UDF behavioral sweep is reported.

\begin{table}[!t]
\centering
\scriptsize
\setlength{\tabcolsep}{3pt}
\begin{adjustbox}{max width=\linewidth}
\begin{tabular}{lrrrr}
\toprule
channel & \#runs & wall-h & GPU-h(TP2) & tokens(M) \\
\midrule
AR-Base (24 policies) & 24 & 2.30 & 4.61 & 123.8 \\
AR-RL (24 policies) & 24 & 2.05 & 4.11 & 122.1 \\
Expanded base (dedup) & 8 & 225.93 & 451.87 & 65.9 \\
\midrule
\multicolumn{5}{l}{\textit{Expanded breakdown by controller family}} \\
\quad EntTree ($b{=}4$) & 1 & 42.11 & 84.21 & 9.5 \\
\quad MCMC ($b{=}4$, 3 samplers) & 3 & 63.58 & 127.17 & 17.8 \\
\quad MCTS (16 iters, 2 samplers) & 2 & 63.38 & 126.76 & 19.6 \\
\quad Tree ($b\in\{4,8\}$) & 2 & 56.86 & 113.73 & 18.9 \\
\midrule
\textbf{Anchor-cell total} & \textbf{56} & \textbf{230.29} & \textbf{460.58} & \textbf{311.8} \\
\bottomrule
\end{tabular}
\end{adjustbox}
\caption{UDF sweep compute cost on the Qwen2.5-7B / Math500 anchor cell, deduplicated at the generation-run level (the same 16-sample pool is reused across budgets $\{1,2,4,8,16\}$ and metrics $\{$pass@$k$, best-of-$n$, majority-vote, first-finish$\}$).  GPU-hours assume vLLM with tensor-parallel $=2$ on H100s.  AR sweeps cover the 23 base-shared policies plus the RL default \texttt{topp0.95\_t1.0}.  Expanded controllers are the cross-family entries used in the recovery near-match search (Section~\ref{sec:recovery}, Fig.~\ref{fig:recovery}).}
\label{tab:udf-cost}
\end{table}

\FloatBarrier
\subsection{BOPTR supplements}
\label{app:boptr-supp}

\subsubsection{Anchor matching set}

\begin{figure}[t]
    \centering
    \includegraphics[width=0.7\linewidth]{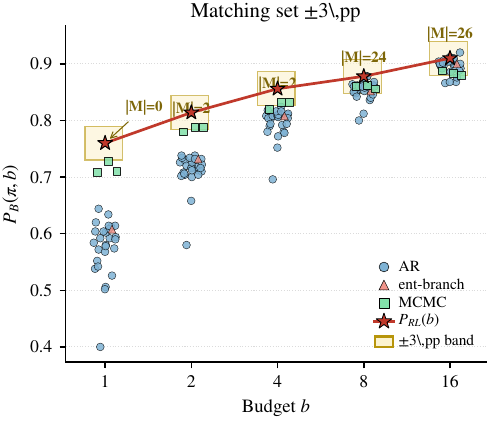}
    \caption{Anchor cell (Qwen2.5-7B/Math500): every RL operating point \(P_{\rl}(b)\) is matched within \(\pm 3\) pp by multiple Base+\udf{} \((\pi,N)\) points.  The matching set is rich, so the rule's job is to pick a budget-monotone path, not to find scarce structure.  Supplement to Section~\ref{sec:boptr-setup}.}
    \label{fig:boptr-matching}
\end{figure}

\subsubsection{Benchmark-regime \(\beta\) visualization}

\begin{figure}[t]
    \centering
    \includegraphics[width=0.7\linewidth]{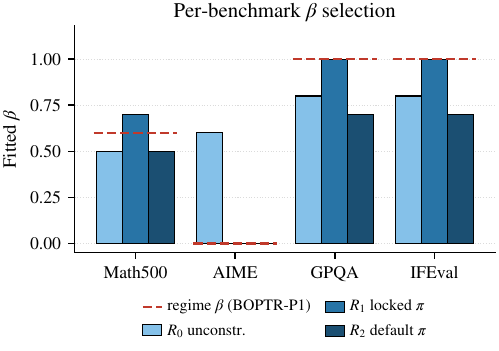}
    \caption{Per-benchmark R0/R1/R2 \(\beta\) fits.  The benchmark splits into three regimes --- math (\(\beta\!=\!0.6\)), nonmath OOD (\(\beta\!=\!1.0\)), and a third, budget-insensitive regime for which \(\beta\!=\!0\) fits best --- rather than a universal exponent.  We label this third regime ``floor'' descriptively, not as a claim about a fixed accuracy floor; Table~\ref{tab:reb-beta} examines the \(\beta\!=\!0\) assignment on these benchmarks directly.  Visualization of main Table~\ref{tab:regime-horizontal}A.}
    \label{fig:beta-robustness}
\end{figure}

\subsubsection{Horizontal-transfer bar chart}

\begin{figure}[t]
    \centering
    \includegraphics[width=0.7\linewidth]{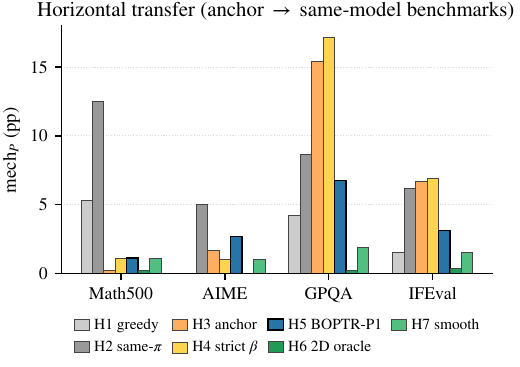}
    \caption{Horizontal-transfer baseline comparison (anchor=Qwen2.5-7B/Math500 \(\to\) other benchmarks).  Bars are row-mean mech\(_P\) over four benchmarks; H5 (regime \(\beta\)) is the lowest non-oracle.  Visualization of main Table~\ref{tab:regime-horizontal}B.}
    \label{fig:horizontal}
\end{figure}

\subsubsection{Vertical-transfer heatmap}

\begin{figure}[t]
    \centering
    \includegraphics[width=0.7\linewidth]{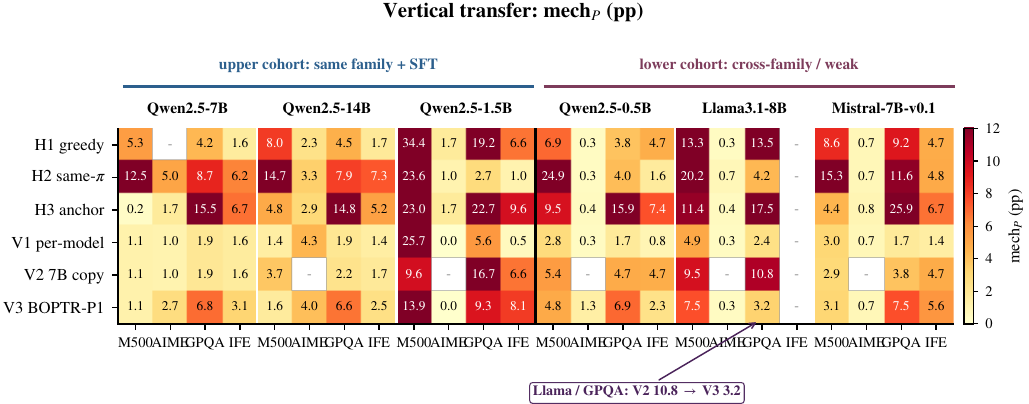}
    \caption{Vertical-transfer heatmap (7 models \(\times\) 4 benchmarks).  V2 (anchor path copy, no target-RL information) is competitive in the upper same Hard RL-training split cohort (including Qwen2.5-Math-7B) but breaks down on the cross-family lower cohort.  Visualization of main Table~\ref{tab:vertical-combined}A.}
    \label{fig:vertical}
\end{figure}

\subsubsection{Formula-same, parameters-different}

\begin{figure}[t]
    \centering
    \includegraphics[width=0.7\linewidth]{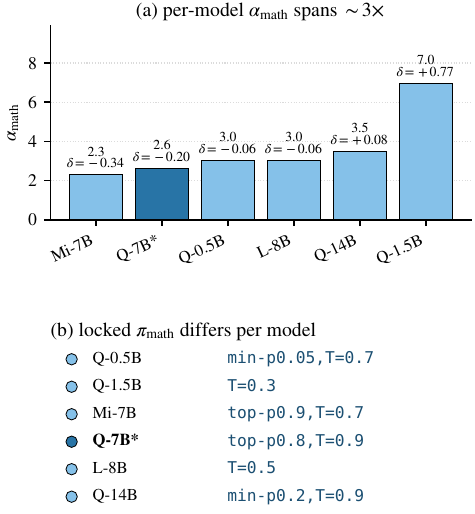}
    \caption{Parameters differ but the formula is universal: \(\alpha_{\text{math}}\) spans 3.0\(\times\) across the seven models, and each model picks its own R1-locked \(\pi_{\text{math}}\).  BOPTR absorbs all of this into the single scalar \(\delta_m\).}
    \label{fig:params}
\end{figure}

Across the seven models the BOPTR-P1 rule keeps its structural form (regime \(\beta\) shared, selector procedure shared) while two per-model knobs absorb individual differences: a continuous scale \(\alpha_{\text{math}}\) and a discrete locked policy \(\pi_{\text{math}}\) (Table~\ref{tab:formula-same}).  The scale spans \(3.0\times\) (2.30--6.96), and the locked \(\pi_{\text{math}}\) varies across models (\texttt{topp}, \texttt{minp}, and pure temperature all appear).  This is the empirical content of ``formula-same, parameters-different'': the rule transfers, but the per-cell tuple \((\alpha,\pi)\) does not, and trying to copy any single anchor policy across models would lose the \(\pi\) component every time.  Figure~\ref{fig:params} visualizes the same fact.

\begin{table}[t]
\centering
\small
\setlength{\tabcolsep}{3pt}
\begin{tabular}{lcc}
\toprule
model & \(\alpha_{\text{math}}\) & \(\pi_{\text{math}}\) (R1 locked) \\
\midrule
Qwen2.5-7B & 2.64 & \texttt{topp0.8\_t0.9} \\
Qwen2.5-14B & 3.48 & \texttt{minp0.2\_t0.9} \\
Qwen2.5-1.5B & 6.96 & \texttt{temp\_0.3} \\
Qwen2.5-0.5B & 3.03 & \texttt{minp0.05\_t0.7} \\
Qwen2.5-Math-7B & 2.64 & \texttt{minp0.05\_t0.7} \\
Llama-3.1-8B & 3.03 & \texttt{temp\_0.5} \\
Mistral-7B-v0.1 & 2.30 & \texttt{topp0.9\_t0.7} \\
\bottomrule
\end{tabular}
\caption{Formula-same, parameters-different.  All seven models share the same rule with regime \(\beta\) fixed; only \(\alpha\) and the locked \(\pi\) change per cell.  \(\alpha_{\text{math}}\) spans 3.0\(\times\); \(\pi_{\text{math}}\) varies across models --- no transferable single policy, only a transferable selection procedure.  \(\alpha_{\text{nonmath}}\!=\!1.0\) for every model by construction (\(\beta\!=\!1\) leaves \(\alpha\) no degrees of freedom).}
\label{tab:formula-same}
\end{table}

\subsubsection{Full BOPTR-P1 per-cell results (23 cells)}
\label{app:boptr-percell}

Table~\ref{tab:boptr-percell} reports the per-cell BOPTR-P1 rule parameters (regime \(\beta\), anchor-calibrated \(\alpha\), selected \(\widehat{\pi}\)) and the resulting mech\(_P\) per cell, with the cell-fit reference column.  The anchor-vs-cell-fit gap is the transfer cost of replacing per-cell \(\alpha\) fits with the 1-D \(\delta_m\) calibration.

The gap column makes the transfer cost concrete.  Most cells transfer at zero cost: 16/23 cells show gap \(=\!+0.00\), meaning the anchor-calibrated \(\alpha\) already matches the per-cell fit.  Three cells absorb a moderate cost (qwen25\_7b/aime \(+1.67\), qwen25\_1.5b/gpqa \(+1.41\), mistral7b\_v01/ifeval \(+2.11\)), and the two outliers --- qwen25\_1.5b/ifeval (\(+6.10\)) and mistral7b\_v01/gpqa (\(-5.45\), where the rule \emph{beats} the cell fit) --- both involve the smallest two models on OOD-nonmath benchmarks, where the cell-fit \(\alpha\) is itself unstable.  The negative gap for mistral7b\_v01/gpqa is a reminder that the per-cell oracle is not a strict lower bound when the cell fit overfits five budgets with two free parameters.

\begin{table}[t]
\centering
\small
\setlength{\tabcolsep}{4pt}
\begin{tabular}{llcccrrrr}
\toprule
model & benchmark & regime & \(\beta\) & \(\widehat{\alpha}\) & \(\widehat{\pi}\) & mech\(_P^{\mathrm{cell}}\) & mech\(_P^{\mathrm{anchor}}\) & gap \\
\midrule
qwen25\_0.5b & math500 & math    & 0.6 & 3.03 & \texttt{topp0.8\_t0.9}    & 4.80  & 4.80  & +0.00 \\
qwen25\_0.5b & aime    & floor   & 0.0 & 1.82 & \texttt{minp0.1\_t0.9}    & 0.33  & 1.33  & +1.00 \\
qwen25\_0.5b & gpqa    & nonmath & 1.0 & 0.91 & \texttt{topk10\_t0.9}     & 6.87  & 6.87  & +0.00 \\
qwen25\_0.5b & ifeval  & nonmath & 1.0 & 0.91 & \texttt{topp0.9\_t0.9}    & 2.33  & 2.33  & +0.00 \\
qwen25\_1.5b & math500 & math    & 0.6 & 6.96 & \texttt{topk20\_t0.9}     & 13.92 & 13.92 & +0.00 \\
qwen25\_1.5b & aime    & floor   & 0.0 & 4.19 & \texttt{topp0.9\_t0.7}    & 1.67  & 0.00  & \(-\)1.67 \\
qwen25\_1.5b & gpqa    & nonmath & 1.0 & 2.10 & \texttt{topk20\_t0.9}     & 7.88  & 9.29  & +1.41 \\
qwen25\_1.5b & ifeval  & nonmath & 1.0 & 2.10 & \texttt{topk40\_t0.7}     & 1.96  & 8.06  & +6.10 \\
qwen25\_7b   & math500 & math    & 0.6 & 2.64 & \texttt{topp0.8\_t0.9}    & 1.12  & 1.12  & +0.00 \\
qwen25\_7b   & aime    & floor   & 0.0 & 1.59 & \texttt{topk20\_t0.7}     & 1.00  & 2.67  & +1.67 \\
qwen25\_7b   & gpqa    & nonmath & 1.0 & 0.79 & \texttt{minp0.05\_t0.9}   & 6.77  & 6.77  & +0.00 \\
qwen25\_7b   & ifeval  & nonmath & 1.0 & 0.79 & \texttt{topk10\_t0.7}     & 3.11  & 3.11  & +0.00 \\
qwen25\_14b  & math500 & math    & 0.6 & 3.48 & \texttt{topp0.8\_t0.9}    & 1.64  & 1.64  & +0.00 \\
qwen25\_14b  & aime    & floor   & 0.0 & 2.10 & \texttt{topp0.9\_t0.9}    & 4.33  & 4.00  & \(-\)0.33 \\
qwen25\_14b  & gpqa    & nonmath & 1.0 & 1.05 & \texttt{topk20\_t0.7}     & 6.57  & 6.57  & +0.00 \\
qwen25\_14b  & ifeval  & nonmath & 1.0 & 1.05 & \texttt{topk40\_t0.7}     & 2.48  & 2.48  & +0.00 \\
llama31\_8b  & math500 & math    & 0.6 & 3.03 & \texttt{minp0.1\_t0.7}    & 7.48  & 7.48  & +0.00 \\
llama31\_8b  & aime    & floor   & 0.0 & 1.82 & \texttt{topp0.95\_t1.0}   & 0.33  & 0.33  & +0.00 \\
llama31\_8b  & gpqa    & nonmath & 1.0 & 0.91 & \texttt{topk20\_t0.9}     & 3.23  & 3.23  & +0.00 \\
mistral7b\_v01 & math500 & math    & 0.6 & 2.30 & \texttt{topp0.95\_t0.7}  & 3.08  & 3.08  & +0.00 \\
mistral7b\_v01 & aime    & floor   & 0.0 & 1.38 & \texttt{topp0.95\_t1.0}  & 0.67  & 0.67  & +0.00 \\
mistral7b\_v01 & gpqa    & nonmath & 1.0 & 0.69 & \texttt{topp0.95\_t0.9}  & 12.93 & 7.47  & \(-\)5.45 \\
mistral7b\_v01 & ifeval  & nonmath & 1.0 & 0.69 & \texttt{topp0.95\_t0.7}  & 3.48  & 5.58  & +2.11 \\
\bottomrule
\end{tabular}
\caption{Full BOPTR-P1 P-channel per-cell results (mean mech\(_P\) over budgets, pp).  ``cell'' = per-cell \(\alpha\) fit (oracle upper bound on the rule); ``anchor'' = anchor-calibrated \(\alpha\) (\(\mu_g+\delta_m\), 1-D model offset).  Transfer gap is anchor\(-\)cell.  ifeval missing for llama31\_8b.}
\label{tab:boptr-percell}
\end{table}

\subsubsection{Ablation study: selector, \(\delta_m\), and \(\widehat{N}(b)\)}
\label{app:boptr-ablation}

We ablate the three structural ingredients of BOPTR-P1 (Table~\ref{tab:boptr-ablations}): the behavior-coordinate \(\pi\) selector (A1 replaces \(\widehat{\pi}\) with the cell's RL default), the per-model \(\delta_m\) offset (A2 sets \(\delta_m=0\)), and the predicted \(\widehat{N}(b)\) (A3 fixes \(N=b\)).  Both 1-D and 2-D oracles bound the rule from below.

The largest single drop is A1 (\(+2.76\) pp on the 23-cell mean), confirming that the behavior-coordinate selector is the load-bearing ingredient and that copying the RL default policy back into a base evaluation undershoots its effective sample efficiency.  A3 (no \(\widehat{N}(b)\) rule, \(+1.17\) pp) shows the budget exponent does real work on math (\(+5.5\) pp on math regime alone) but adds nothing on nonmath/floor regimes whose \(\beta\!\in\!\{1,0\}\) makes \(\widehat{N}(b)\) a no-op anyway.  A2 (no \(\delta_m\)) is the most subtle: it is slightly better than MAIN on the 23-cell flat mean (\(-0.31\) pp), but its real role is the cross-family rescue in Table~\ref{tab:vertical-combined}B where Llama drops from 10.16 to 3.68 pp.  This is consistent with the \(\delta_m\) story: a per-model offset matters precisely where a single anchor's scale is wrong.

\begin{table}[t]
\centering
\small
\setlength{\tabcolsep}{2pt}
\begin{adjustbox}{max width=\linewidth}
\begin{tabular}{lcccc}
\toprule
variant & math & nonmath & floor & 23-cell mean \\
\midrule
\textbf{MAIN (BOPTR-P1)} & \textbf{5.34} & \textbf{5.61} & \textbf{1.50} & \textbf{4.47} \\
A1 no \(\widehat{\pi}\) selector & 12.11 & 7.20 & 2.39 & 7.23 \\
A2 no \(\delta_m\) offset        & 4.78 & 5.24 & 1.56 & 4.16 \\
A3 no \(\widehat{N}(b)\) rule    & 10.85 & 5.24 & 1.17 & 5.64 \\
\midrule
ORACLE-2D per-\(b\) (\(\pi,N\)) & 1.59 & 0.39 & 0.11 & 0.63 \\
ORACLE-1D per-\(b\) (\(\pi\), \(N\!=\!b\)) & 7.18 & 0.54 & 0.17 & 2.17 \\
\bottomrule
\end{tabular}
\end{adjustbox}
\caption{BOPTR-P1 ablation, regime-aggregated mean mech\(_P\) (pp).  A1 (no selector) is the largest drop on math (\(+6.8\) pp).  A3 (no compute rule) hurts most on math (\(+5.5\) pp).  A2 is slightly \emph{better} than MAIN at the 23-cell mean (\(-0.31\) pp); its real role (Table~\ref{tab:vertical-combined}B) is closing the cross-family gap in the lower half.}
\label{tab:boptr-ablations}
\end{table}

\subsubsection{Q channel: rule + recoverability}
\label{app:boptr-q}

Table~\ref{tab:boptr-q} reports the Q-channel (self-consistency) BOPTR rule and the cell-level recoverability flag.  The Q channel uses \(\widehat{\pi}^Q=\arg\max_\pi[Q_B(\pi,16)-0.5\,(P_B-Q_B)-0.01\,C]\) and \(\beta^Q\!=\!0\) outside GPQA (which is the only regime where Q scales with budget).  19/23 cells are recoverable within 0.05 of max \(Q_{\rl}\); the four unrecoverable cells are all (model, Math500) where RL has discovered a sharper distribution than any base policy at any budget --- this is the SC-side analog of the same-policy gap in Section~\ref{sec:recovery}.

The four unrecoverable Math500 cells span 0.5B/1.5B/14B Qwen and Llama-3.1-8B; the 7B and Mistral-7B Math500 cells \emph{are} recoverable, so unrecoverability is not monotone in capacity.  All four use the same \(\widehat{\alpha}^Q\!=\!16\) (the budget cap), reflecting that the Q-channel selector pushes to the maximum allowed sample budget when no base policy reaches RL's SC concentration.  Floor regime is absent here because AIME does not produce a stable Q-channel signal at our scale; the table therefore covers Math500 (math) and GPQA (nonmath) only.

\begin{table}[t]
\centering
\small
\setlength{\tabcolsep}{2pt}
\begin{adjustbox}{max width=\linewidth}
\begin{tabular}{llcccr}
\toprule
model & benchmark & recov & \(\beta^Q\) & \(\widehat{\alpha}^Q\) & mech\(_Q\) (pp) \\
\midrule
qwen25\_0.5b & math500 & \(\times\) & 0.0 & 16.0 & 5.32 \\
qwen25\_0.5b & gpqa    & \(\checkmark\) & 0.6 & 0.44 & 2.22 \\
qwen25\_1.5b & math500 & \(\times\) & 0.0 & 16.0 & 11.08 \\
qwen25\_1.5b & gpqa    & \(\checkmark\) & 0.6 & 1.74 & 6.16 \\
qwen25\_7b   & math500 & \(\checkmark\) & 0.0 & 16.0 & 1.64 \\
qwen25\_7b   & gpqa    & \(\checkmark\) & 0.6 & 3.48 & 3.13 \\
qwen25\_14b  & math500 & \(\times\) & 0.0 & 16.0 & 4.20 \\
qwen25\_14b  & gpqa    & \(\checkmark\) & 0.6 & 3.48 & 2.53 \\
llama31\_8b  & math500 & \(\times\) & 0.0 & 16.0 & 4.72 \\
llama31\_8b  & gpqa    & \(\checkmark\) & 0.6 & 1.74 & 4.44 \\
mistral7b\_v01 & math500 & \(\checkmark\) & 0.0 & 16.0 & 2.40 \\
mistral7b\_v01 & gpqa    & \(\checkmark\) & 0.6 & 1.74 & 4.04 \\
\midrule
\multicolumn{6}{l}{\textit{Recoverability:} 19/23 cells (83\%).} \\
\multicolumn{6}{l}{\textit{Unrecoverable:} 4 math500 cells.} \\
\bottomrule
\end{tabular}
\end{adjustbox}
\caption{BOPTR Q-channel rule + recoverability.  Selected cells; full 23-cell results in supplementary CSV.  Mean mech\(_Q\) (recoverable only): math 2.02 pp, nonmath 2.76 pp, floor 0.44 pp.}
\label{tab:boptr-q}
\end{table}

\subsubsection{V4: a linear base-only ridge for \(\alpha_{\text{math}}\) is underpowered at \(n{=}6\)}
\label{app:boptr-v4}

\begin{figure}[t]
    \centering
    \includegraphics[width=0.7\linewidth]{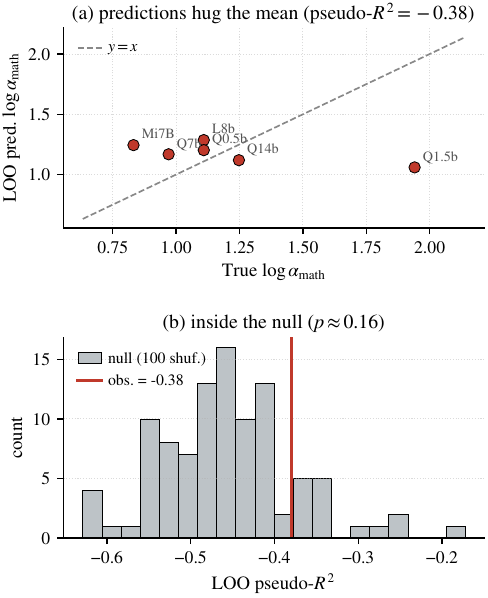}
    \caption{A single linear ridge on base features predicting \(\log\alpha_{\text{math}}\) sits inside the \(y\)-shuffled null (empirical \(p\!\approx\!0.21\)) at \(n\!=\!6\) models.  The main text uses a one-cell calibration baseline (V3) that estimates \(\delta_m\) from each model's Math500 Base/RL pair; a structured base-only predictor (V5b) recovers most of the same offset without target-model RL calibration (Table~\ref{tab:reb-vertical}B).}
    \label{fig:v4-negative}
\end{figure}

A natural follow-up is whether the per-model offset \(\delta_m\) itself can be predicted from base-only statistics --- i.e.\ without any RL data at all.  Table~\ref{tab:v4-negative} and Figure~\ref{fig:v4-negative} report a leave-one-model-out ridge of \(\log\alpha_{\text{math}}\) on 24 cross-benchmark base features (six per benchmark) plus \(\log(\text{model\_size})\).  With only 6 models, the top single feature (\texttt{gpqa\_\_P0\_b16}) achieves LOO pseudo-\(R^2=-0.38\) with empirical \(p\!\approx\!0.21\) over 100 \(y\)-shuffles --- well inside the null.  Multi-feature ridge does not help (\(k\!=\!1\) gives LOO pseudo-\(R^2\!=\!-0.43\); \(k\!=\!2\) gives \(-0.42\); \(k\!=\!3\) gives \(-0.43\); ridge \(\lambda\!=\!1.0\)).  We treat this as a genuine limit of \emph{this} predictor, not noise: a single linear ridge over one feature family does not identify \(\delta_m\) at \(n\!=\!6\).  It does not follow that \(\delta_m\) is unpredictable from base statistics.  A typicality-gated multi-signal predictor (V5b, Table~\ref{tab:reb-vertical}B and \S\ref{sec:boptr-vertical}) recovers most of the offset from base-side features on the same cohort, and the RL-information frontier below reaches the same conclusion by a different route (Appendix~\ref{app:rl-info-frontier}).

The feature ranking is itself informative.  All five top features are accuracy- or entropy-style metrics on \emph{other} benchmarks (GPQA and Math500 base statistics), not the within-benchmark Math500 base curve --- consistent with the BOPTR claim that \(\delta_m\) is a cross-benchmark property of the model, not a per-cell artifact.  The negative \(r\) for \texttt{gpqa\_\_P0\_b16} means models with stronger base GPQA accuracy tend to need \emph{smaller} \(\alpha_{\text{math}}\), which is mechanistically sensible (a model that already concentrates on the right answers at \(b\!=\!16\) does not benefit from large effective sample budgets).  But with \(n\!=\!6\) the in-sample \(|r|\!=\!0.79\) does not survive LOO for this linear probe.  Turning these base-side signals into a usable estimate of \(\delta_m\) requires a richer, gated predictor rather than a larger linear ridge, which is what V5b supplies.

\begin{table}[t]
\centering
\small
\setlength{\tabcolsep}{3pt}
\begin{adjustbox}{max width=\linewidth}
\begin{tabular}{lcc}
\toprule
feature (top by \(|r|\)) & Pearson \(r\) & LOO pseudo-\(R^2\) \\
\midrule
\texttt{gpqa\_\_P0\_b16} & \(-0.79\) & \(-0.38\) \\
\texttt{gpqa\_\_Pmax\_b16} & \(-0.77\) & \(-0.39\) \\
\texttt{gpqa\_\_Dp} & \(+0.64\) & \(-0.46\) \\
\texttt{math500\_\_Dp} & \(+0.59\) & \(-0.44\) \\
\texttt{math500\_\_Hp} & \(+0.51\) & \(-0.44\) \\
\midrule
\multicolumn{3}{l}{\textit{Null (100 \(y\)-shuffles, top-1):}} \\
\multicolumn{3}{l}{5\%/50\%/95\% \(=\) \(-0.60/{-0.46}/{-0.32}\); \(p\!\approx\!0.21\)} \\
\bottomrule
\end{tabular}
\end{adjustbox}
\caption{Base-only linear LOOCV ridge of \(\log\alpha_{\text{math}}\) (\(n\!=\!6\)).  In-sample \(|r|\) reaches 0.79 but every LOO pseudo-\(R^2\) is negative and the top score sits inside the \(y\)-shuffle null: this single-family linear probe is underpowered at \(n\!=\!6\).  A gated multi-signal predictor (V5b) succeeds where it fails.}
\label{tab:v4-negative}
\end{table}

\subsubsection{RL-information budget frontier}
\label{app:rl-info-frontier}

Table~\ref{tab:rl-info-frontier} sweeps how much target-RL information a predictor of post-RL behavior actually needs, holding the regime exponent \(\beta\) and the behavior-coordinate selector fixed.  We organize predictors by the number of RL models they touch.  The \emph{Zero-RL} tier uses no RL model at all: features are read off the base evaluated on the RL-training distribution (\texttt{base@rltrain}), with $u=\sqrt{(1-P_0)\cdot\mathrm{headroom}}$ a sampling-headroom prior and $N^\star$ the saturation knee (the smallest budget within $\tau$ of the per-cell ceiling, $\log_2$-interpolated; main $\tau{=}0.02$).  The \emph{one-anchor} tier adds a single calibration cell (Qwen2.5-7B/Math500): B0 copies the anchor scale ($\mu_{g}$, $\delta_m{=}0$), V2 copies the anchor $(\pi,N)$ path directly, and Z1 blends the Zero-RL prior with the anchor by a mixing weight $\lambda$.  V3 (seven per-model $\delta_m$ fits from each model's Math500 Base/RL pair) and V1 (per-cell oracle) are reference points.

The frontier is flat above one anchor: Z1-$N^\star$ reaches 4.08~pp with a single RL cell, matching or beating the seven-model V3 (4.47~pp), and even the no-RL Zero-RL prior reaches 5.13~pp.  The information needed to predict post-RL behavior compresses to at most one anchor cell, and approximately to zero.  As a base-only probe, the \texttt{base@rltrain} saturation rate $P_{\max}@16$ correlates with $\log\alpha_{\text{math}}$ at Pearson $r=-0.96$ across the seven models (OLS $\log\alpha = 3.07 - 2.78\,x$).

\begin{table}[H]
\centering
\small
\setlength{\tabcolsep}{4pt}
\begin{tabular}{@{}lccc@{}}
\toprule
rule & \#RL & \makecell{\texttt{base@}\\\texttt{rltrain}} & \makecell{mech\(_P\)\\(pp)} \\
\midrule
\multicolumn{4}{@{}l}{\textit{Zero-RL: \texttt{base@rltrain} features only}} \\
Z0-\(u\) (log2 prior) & 0 & \checkmark & 5.13 \\
Z0-\(N^\star\) (\(\tau{=}0.02\), main) & 0 & \checkmark & 6.13 \\
\midrule
\multicolumn{4}{@{}l}{\textit{One-anchor: \(+\) single Qwen2.5-7B/Math500 RL cell}} \\
B0 (anchor \(\mu\), \(\delta_m{=}0\)) & 1 & \(\times\) & 4.34 \\
V2 anchor-path copy & 1 & \(\times\) & 4.95 \\
Z1-\(u\) (\(\lambda{=}1.0\)) & 1 & \checkmark & 4.32 \\
Z1-\(N^\star\) (\(\lambda{=}1.0\)) & 1 & \checkmark & \textbf{4.08} \\
\midrule
\multicolumn{4}{@{}l}{\textit{Per-model / oracle (reference)}} \\
V3 per-model \(\delta_m\) (Math500) & 7 & \(\times\) & 4.47 \\
V1 per-cell oracle & oracle & \(\times\) & 2.77 \\
\bottomrule
\end{tabular}
\caption{RL-information budget frontier.  Mean mech\(_P\) (pp, over the 27 paired Base/RL cells; \texttt{ifeval} missing for Llama-3.1-8B) as a function of how much target-RL information the predictor uses.  \#RL counts the RL models touched; the \texttt{base@rltrain} column marks whether base-on-RL-distribution features are used.  Z1-\(N^\star\) (one anchor) matches the seven-model V3, and the Zero-RL prior (no RL model) is within \(\sim 1\) pp, so the predictor's reliance on RL data compresses to at most one anchor cell.}
\label{tab:rl-info-frontier}
\end{table}

\subsection{Extended tables from post-submission experiments}
\label{app:extended-tables}

These tables report additional experiments as increments to
Table~\ref{tab:regime-horizontal} and Table~\ref{tab:vertical-combined}:
each table reproduces the original rows and columns verbatim and appends new ones.
Table~\ref{tab:reb-horizontal} extends Table~\ref{tab:regime-horizontal} with a-priori
regime labels for four held-out benchmarks, per-cell bootstrap confidence intervals,
a base-only calibration variant, a decoding-seed replication, a budget extension to
\(b>16\), and held-out benchmark columns.
Table~\ref{tab:reb-vertical} extends Table~\ref{tab:vertical-combined} with per-model
V3 rows, the base-only variant V5b, two additional RL recipes, and a cross-family
model; the held-out model\(\times\)benchmark grid, now complete, appears in the
main text (Table~\ref{tab:coverage-beyond}).
Table~\ref{tab:reb-beta} addresses the \(\beta{=}0\) (floor) assignment directly.
No frozen quantity is refit for any number on these pages: \(\mu_g\), \(\beta_g\),
\(\delta_m\), the regime map, the policy selector, and the budget grid retain their
main-text values.
Unless stated otherwise, errors are mean \(|\widehat{P}_{\rl}-P_{\rl}|\) in percentage
points (pp) over \(b\in\{1,2,4,8,16\}\).

\paragraph{The base-only variant V5b.}
The main text estimates each model's offset \(\delta_m\) from its Math500 Base/RL
pair (V3) and reports a base-only prediction of the same offset
(V5b, \S\ref{sec:boptr-vertical}); we give the construction of \(\hat{\delta}_m\) here.
V5b replaces \(\delta_m\) with a prediction \(\hat{\delta}_m\) computed from
base-side features only: a hard typicality gate chooses between a tail-consistency
estimate and a ridge-regression estimate, fit leave-one-model-out so that the target
model contributes no RL data; all other frozen quantities are unchanged.
On the anchor, \(\hat{\delta}=-0.2184\) versus the calibrated \(\delta=-0.1980\)
yields the same \(\widehat{N}\) sequence after log-2 rounding, hence a bit-identical
operating path and identical errors --- this identity is why several V5b entries
below repeat H5/V3 values exactly rather than approximately.

\paragraph{Readout-consistency caliber.}
The RL side was re-sampled for the large-budget extension and checked against the
submission pool on shared (policy, budget) points: mean \(|\Delta|\) = 0.57 / 1.06 /
0.32\,pp on Math500 / AIME / IFEval (within the pre-set 2\,pp criterion), but
3.64\,pp on GPQA with a systematic \(+2.1\)\,pp signed offset concentrated in
high-temperature policies (up to \(+12.12\)\,pp at temperature 1.0, \(b{=}1\)).
GPQA comparisons that cross sampling batches therefore carry a \(\pm\)3--4\,pp
caliber; within-batch comparisons are unaffected.

% ------------------------------------------------------------------ Table R1
\begin{table}[!t]
\scriptsize
\setlength{\tabcolsep}{2pt}
\renewcommand{\arraystretch}{0.96}

% --- (A) compact table on the left, its note alongside -----------------
\begin{tabular}{@{}p{0.315\textwidth}@{\hspace{\columnsep}}p{0.64\textwidth}@{}}
\begin{minipage}[t]{\linewidth}
\vspace{0pt}
\raggedright
\begin{tabular}{@{}lcccc@{}}
\toprule
benchmark & regime & R0 \(\beta\) & R1 \(\beta\) & R2 \(\beta\) \\
\midrule
Math500 & math & 0.50 & 0.70 & 0.50 \\
AIME & floor & 0.60 & 0.00 & 0.00 \\
GPQA & nonmath & 0.80 & 1.00 & 0.70 \\
IFEval & nonmath & 0.80 & 1.00 & 0.70 \\
\midrule
AMC23 & math\(^{\mathrm{a}}\) & --- & --- & --- \\
Minerva & math\(^{\mathrm{a}}\) & --- & --- & --- \\
OlympiadBench & math\(^{\mathrm{a}}\) & --- & --- & --- \\
TinyMMLU & nonmath\(^{\mathrm{a}}\) & --- & --- & --- \\
\bottomrule
\end{tabular}
\end{minipage}
&
\begin{minipage}[t]{\linewidth}
\vspace{0pt}
\raggedright
\panelnote{\textbf{(A)} Regime discovery. R0 allows \(\pi\) to vary across budgets,
R1 locks one \(\pi\), and R2 uses the default \(\pi\).
Added rows: \(^{\mathrm{a}}\)regime assigned a priori (pre-registered before any
evaluation).
Under the no-refit protocol no \(\beta\) is fit for these benchmarks, so their
R0--R2 columns are empty by design; transfer quality is instead tested directly in
(B) and in Table~\ref{tab:reb-vertical}(C).}
\end{minipage}
\end{tabular}

\vspace{1.4ex}

% --- (B) full-width table, note underneath ----------------------------
\begin{adjustbox}{max width=\textwidth}
\begin{tabular}{@{}lcccccccccc@{}}
\toprule
& \multicolumn{5}{c}{submission grid} & \multicolumn{5}{c@{}}{held-out, frozen rule (new)} \\
\cmidrule(lr){2-6}\cmidrule(l){7-11}
baseline & Math500 & AIME & GPQA & IFEval & mean & AMC23 & Minerva & Olymp. & TinyMMLU & mean \\
\midrule
H1 base greedy &
\makecell[t]{5.32\\[-0.7ex]{\tiny[3.84, 7.48]}} & --- &
\makecell[t]{4.24\\[-0.7ex]{\tiny[1.62, 8.69]}} &
\makecell[t]{1.55\\[-0.7ex]{\tiny[0.70, 4.55]}} &
\makecell[t]{3.70\\[-0.7ex]{\tiny[2.78, 5.74]}} & --- & --- & --- & --- & --- \\
H2 base same-\(\pi\) &
\makecell[t]{12.52\\[-0.7ex]{\tiny[10.24, 14.84]}} &
\makecell[t]{5.00\\[-0.7ex]{\tiny[0.67, 10.33]}} &
\makecell[t]{8.69\\[-0.7ex]{\tiny[5.56, 12.83]}} &
\makecell[t]{6.21\\[-0.7ex]{\tiny[4.73, 7.99]}} &
\makecell[t]{8.10\\[-0.7ex]{\tiny[6.64, 10.00]}} & --- & --- & --- & --- & --- \\
H3 anchor per-\(b\) copy &
\makecell[t]{0.20\\[-0.7ex]{\tiny[0.40, 2.04]}} &
\makecell[t]{1.67\\[-0.7ex]{\tiny[0.00, 5.83]}} &
\makecell[t]{15.46\\[-0.7ex]{\tiny[12.12, 18.99]}} &
\makecell[t]{6.70\\[-0.7ex]{\tiny[4.99, 8.96]}} &
\makecell[t]{6.01\\[-0.7ex]{\tiny[5.31, 7.82]}} & --- & --- & --- & --- & --- \\
H4 strict BOPTR (\(\beta\!=\!0.6\)) &
\makecell[t]{1.12\\[-0.7ex]{\tiny[0.76, 2.52]}} &
\makecell[t]{1.00\\[-0.7ex]{\tiny[0.33, 6.67]}} &
\makecell[t]{17.17\\[-0.7ex]{\tiny[11.62, 22.63]}} &
\makecell[t]{6.95\\[-0.7ex]{\tiny[4.14, 9.80]}} &
\makecell[t]{6.56\\[-0.7ex]{\tiny[5.39, 8.86]}} & --- & --- & --- & --- & --- \\
\textbf{H5 BOPTR-P1 (regime \(\beta\))} &
\makecell[t]{\textbf{1.12}\\[-0.7ex]{\tiny[0.76, 2.52]}} &
\makecell[t]{\textbf{2.67}\\[-0.7ex]{\tiny[0.00, 8.33]}} &
\makecell[t]{\textbf{6.77}\\[-0.7ex]{\tiny[3.23, 12.22]}} &
\makecell[t]{\textbf{3.11}\\[-0.7ex]{\tiny[1.63, 5.21]}} &
\makecell[t]{\textbf{3.41}\\[-0.7ex]{\tiny[2.32, 5.53]}} &
\textbf{3.00} & \textbf{3.75} & \textbf{2.84} & \textbf{1.60} & \textbf{2.80} \\
\;\;\(\hookrightarrow\) base-only \(\hat{\delta}\) (V5b, new) & 1.12 & 2.67 & 6.77 & 3.11 & 3.41 & 3.00 & 3.75 & 2.84 & 1.60 & 2.80 \\
\;\;\(\hookrightarrow\) 3-seed replication, fixed \(\hat{\pi}^{*}\)\(^{\dagger}\) & 1.96\,{\tiny\(\pm\)0.84} & 3.22\,{\tiny\(\pm\)2.22} & 5.12\,{\tiny\(\pm\)1.50} & 2.05\,{\tiny\(\pm\)0.93} & 3.09\,{\tiny\(\pm\)0.40} & --- & --- & --- & --- & --- \\
\;\;\(\hookrightarrow\) 3-seed replication, full rule\(^{\dagger}\) & 1.61\,{\tiny\(\pm\)0.45} & 3.22\,{\tiny\(\pm\)2.22} & 4.85\,{\tiny\(\pm\)1.76} & 2.59\,{\tiny\(\pm\)0.69} & 3.07\,{\tiny\(\pm\)0.39} & --- & --- & --- & --- & --- \\
\;\;\(\hookrightarrow\) frozen rule at \(b\!>\!16\)\(^{\ddagger}\) & 0.80 & 12.50 & 5.30 & 1.48 & 5.02 & --- & --- & --- & --- & --- \\
H6 2D oracle per-\(b\) &
\makecell[t]{0.20\\[-0.7ex]{\tiny[0.40, 2.04]}} &
\makecell[t]{0.00\\[-0.7ex]{\tiny[0.00, 7.33]}} &
\makecell[t]{0.20\\[-0.7ex]{\tiny[1.11, 5.56]}} &
\makecell[t]{0.33\\[-0.7ex]{\tiny[0.59, 3.11]}} &
\makecell[t]{0.19\\[-0.7ex]{\tiny[1.12, 3.34]}} & --- & --- & --- & --- & --- \\
H7 2D oracle smooth (R1 path) &
\makecell[t]{1.12\\[-0.7ex]{\tiny[0.76, 2.52]}} &
\makecell[t]{1.00\\[-0.7ex]{\tiny[0.00, 7.00]}} &
\makecell[t]{1.92\\[-0.7ex]{\tiny[1.21, 6.87]}} &
\makecell[t]{1.55\\[-0.7ex]{\tiny[0.70, 4.55]}} &
\makecell[t]{1.40\\[-0.7ex]{\tiny[1.31, 3.86]}} & --- & --- & --- & --- & --- \\
\bottomrule
\end{tabular}
\end{adjustbox}

\vspace{0.2ex}

\panelnote{\textbf{(B)} Horizontal transfer. H5 is the strongest non-oracle rule;
all point estimates in the submission columns are unchanged (printed values kept
verbatim; recomputing a mean at full precision can shift its last digit by
\(\le\)0.01).
Brackets are 95\% percentile bootstrap CIs (\(B{=}10^{4}\), question-level
resampling with draws shared across rows, hence paired; each row's
\((\hat{\pi},\widehat{N})\) path held fixed); for scale, the Wilson 95\% half-width
of a single readout at the same \(n\) is
\(\pm\)2.5--3.8 (Math500), \(\pm\)6.0--7.2 (AIME), \(\pm\)3.6--6.9 (GPQA),
\(\pm\)3.9--4.2 (IFEval)\,pp.
Paired differences of row means:
\(\Delta(\text{H5}{-}\text{H2})=-4.69\;[-6.18,-2.75]\),
\(\Delta(\text{H5}{-}\text{H3})=-2.59\;[-4.09,-1.26]\), and
\(\Delta(\text{H5}{-}\text{H4})=-3.15\;[-5.00,-1.80]\) all exclude 0;
\(\Delta(\text{H5}{-}\text{H1})=-0.04\;[-2.69,+2.50]\) indicates parity (H1
defines no AIME cell, so its mean covers three benchmarks).
H6/H7 read per-cell RL information and serve as reference ceilings, not
comparators; descriptively, \(\Delta(\text{H5}{-}\text{H7})=+2.02\;[-0.35,+3.24]\).
Because the statistic is an absolute error, its bootstrap distribution is
upward-biased for near-zero cells, so an interval can lie above its point estimate
(H3 Math500, H6 row); intervals are reported as-is.
The V5b row equals H5 exactly on every column shown (anchor path identity; same
CIs): base-only calibration loses nothing on the anchor.
\(^{\dagger}\)mean\,\(\pm\)\,sd of mech\(_P\) over 3 decoding seeds (42/123/20240;
generation parameters identical, only the seed differs; seed 42 reuses the
submission grids), scored against the RL reference \texttt{top-p 0.95, T=1.0}.
Two readings of the same replication: \emph{fixed \(\hat{\pi}^{*}\)} freezes the
policy at the submission's per-benchmark \(\hat{\pi}^{*}\) (prediction noise only,
no dependence on pool composition; its seed-42 value is arithmetically identical
to the printed cell on all four benchmarks), while \emph{full rule} reruns the
policy selector at each seed before scoring (at seed 42 it also reproduces both
the printed cell and the printed \(\hat{\pi}^{*}\) on all four benchmarks).
Per-seed fixed-\(\hat{\pi}^{*}\) values: Math500 1.12/2.80/1.96, AIME
2.67/1.33/5.67, GPQA 6.77/4.75/3.84, IFEval 3.11/1.66/1.37; the AIME spread
(range 4.33\,pp) equals \(\approx\)2.6 questions at \(n{=}60\).
The mean column averages the four benchmarks within each seed, then reports
mean\,\(\pm\)\,sd over seeds.
Selector pools: AIME contains all 6 submission AR policies (the remaining 10
submission-pool entries are structured decoders outside this runner); the two
readings print identical AIME values --- at seed 123 the selector picks a
different policy whose five-budget mean rounds to the same 1.33.
Math500/GPQA/IFEval rerun pools are complete: IFEval's 24 policies are exactly
the submission pool (its full-rule row is an exact submission-pool
reconstruction); GPQA covers 24 of 34 and Math500 23 of 31 submission entries,
the exclusions being structured decoders outside this runner (plus, for Math500,
\texttt{top-p 0.95, T=1.0}, which exists only on the RL side).
Per-seed full-rule values, scored against the submission RL reference:
Math500 1.12/2.00/1.72, GPQA 6.77/4.44/3.33, IFEval 3.11/2.85/1.81;
restricting seed 42 to these AR pools changes neither its selection nor its
value, so all three seeds are scored on a common pool.
\(^{\ddagger}\)the frozen rule applied beyond its calibration range,
\(b\in\{32,\dots,b_{\max}\}\), \(b_{\max}{=}64\) (Math500/GPQA/IFEval) and 256
(AIME); the AIME error reflects the floor allocation \(\widehat{N}\!\equiv\!2\),
analyzed in Table~\ref{tab:reb-beta}.}

\vspace{0.4ex}

\caption{\footnotesize
Extension of Table~\ref{tab:regime-horizontal} (regime discovery and horizontal
transfer on Qwen2.5-7B; same layout and frozen parameters, no refitting).
(A) adds the four additional benchmarks with a-priori regime labels.
(B) adds bootstrap CIs for every baseline row and row mean (paired
mean-difference CIs in the panel note), a base-only calibration variant, a 3-seed
replication in two readings (policy fixed at \(\hat{\pi}^{*}\) vs.\ full selection
rule), the \(b>16\) extension, and four held-out benchmark columns.
}
\label{tab:reb-horizontal}
\end{table}

% ------------------------------------------------------------------ Table R2
\begin{table}[!t]
\scriptsize
\setlength{\tabcolsep}{2pt}
\renewcommand{\arraystretch}{0.96}

\begin{tabular}{@{}p{\twopanel}@{\hspace{\columnsep}}p{\twopanel}@{}}
\begin{minipage}[t]{\linewidth}
\vspace{0pt}
\raggedright
\begin{adjustbox}{max width=\linewidth}
\begin{tabular}{@{}lccccc@{}}
\toprule
\multicolumn{6}{@{}l}{\textit{Upper: same Hard RL-training split}} \\
baseline & 7B & 14B & 1.5B & Math-7B & mean \\
\midrule
H1 base greedy & 3.71 & 4.16 & 15.47 & 6.63 & 7.49 \\
H2 base same-\(\pi\) & 8.10 & 8.29 & 7.07 & 7.18 & 7.66 \\
H3 anchor per-b copy & 6.01 & 6.92 & 14.24 & 7.58 & 8.69 \\
V1 per-cell oracle & 1.40 & 2.26 & 7.94 & 2.09 & 3.42 \\
V2 anchor path (no target-RL info) & 1.40 & 2.55 & 10.96 & 3.73 & 4.66 \\
\midrule
V3 \(\delta_m\) from Math500 (new row) & 3.41 & 3.67 & 7.82 & 3.55 & 4.61 \\
V5b base-only \(\hat{\delta}_m\) (new) & 3.41 & 3.67 & 5.90 & 3.80 & 4.20 \\
\bottomrule
\end{tabular}
\end{adjustbox}

\vspace{0.45ex}

\begin{adjustbox}{max width=\linewidth}
\begin{tabular}{@{}lcccc@{}}
\toprule
\multicolumn{5}{@{}l}{\textit{Lower: cross-family / weak-capacity}} \\
baseline & 0.5B & Llama & Mistral & mean \\
\midrule
H1 base greedy & 3.93 & 9.06 & 5.79 & 6.26 \\
H2 base same-\(\pi\) & 7.73 & 8.37 & 8.09 & 8.06 \\
H3 anchor per-b copy & 8.32 & 9.79 & 9.44 & 9.18 \\
V1 per-cell oracle & 1.41 & 2.56 & 1.70 & 1.89 \\
V2 anchor path (no target-RL info) & 4.92 & 10.16 & 3.83 & 6.30 \\
\midrule
V3 \(\delta_m\) from Math500 (new row) & 3.83 & 3.68 & 5.12 & 4.21 \\
V5b base-only \(\hat{\delta}_m\) (new) & 3.67 & 3.68 & 5.21 & 4.19 \\
\bottomrule
\end{tabular}
\end{adjustbox}

\vspace{0.45ex}

\begin{adjustbox}{max width=\linewidth}
\begin{tabular}{@{}lccc@{}}
\toprule
\multicolumn{4}{@{}l}{\textit{New models: second RL recipes and a cross-family model}} \\
rule & ORZ-7B & OAT-7B & DS-Math-7B \\
\midrule
V3 \(\delta_m\) from Math500 & 4.87 & 4.48 & 3.28 \\
V5b base-only \(\hat{\delta}_m\) & 4.87 & 4.21 & --- \\
V3 pipeline, \(\delta_m\!:=\!0\) & 4.68 & 4.21 & 3.28 \\
\bottomrule
\end{tabular}
\end{adjustbox}

\panelnote{\textbf{(A)} Direct anchor-path vertical baselines without target-RL
information; each cell is mean mech\(_P\) (pp) over budgets.
Added rows: per-model means for V3 (previously reported only as the 4.44 macro
average) and for the base-only variant V5b.
New-model block: ORZ-7B (Open-Reasoner-Zero) trains from the anchor base, so its
V3 and V5b coincide bitwise; OAT-7B (OAT-Zero) trains from Qwen2.5-Math-7B, and
its V3--V5b difference comes from one budget-bin change at Math500 \(b{=}1\).
DS-Math-7B (DeepSeek-Math-7B) is cross-family and outside the leave-one-model-out
cohort, so the predicted-offset variant V5b is undefined for it (---); it does
have a Math500 Base/RL pair, so V3 is fit from that pair (fitted
\(\delta_m{=}{-}0.06\)). The offset is small enough that the rounded budgets
\(\hat N{=}\mathrm{round}(\alpha b^{\beta})\) match the \(\delta_m\!:=\!0\)
pipeline, so V3 and zero-calibration coincide at 3.28\,pp for this model
(per-benchmark 7.00/0.00/5.45/0.67). The \(\delta_m\!:=\!0\) row is the common
zero-calibration reference across all three new models.}
\end{minipage}
&
\begin{minipage}[t]{\linewidth}
\vspace{0pt}
\raggedright
\begin{adjustbox}{max width=\linewidth}
\begin{tabular}{@{}>{\raggedright\arraybackslash}p{0.30\linewidth}
                >{\raggedright\arraybackslash}p{0.40\linewidth}
                >{\centering\arraybackslash}p{0.24\linewidth}@{}}
\toprule
rule & \makecell{structural\\assumption} & \makecell{overall mean (pp)\\macro / micro} \\
\midrule
V1 per-cell oracle & none (fit \(\alpha,\beta,\pi\) per cell) & 2.77 / 2.77 \\
V2 \(\delta_m\!=\!0\) & formula shared, no offset & 5.36 / 4.95 \\
V3 \(\delta_m\) from Math500 & formula + 1-D offset & 4.44 / 4.47 \\
\textbf{V5b base-only \(\hat{\delta}_m\) (new)} & formula + predicted 1-D offset & \textbf{4.19} / \textbf{4.21} \\
\midrule
\multicolumn{3}{@{}l}{\textit{RL-information frontier (how the offset degrades as RL access shrinks)}} \\
Z1-N\(^{*}\) one RL anchor cell (new) & \(\alpha\) from anchor saturation knee & 4.04 / 4.08 \\
Z0-u no RL at all (new) & \(\alpha\) from base@rltrain fingerprint & 5.08 / 5.13 \\
\bottomrule
\end{tabular}
\end{adjustbox}

\panelnote{\textbf{(B)} Formula-decomposition ablation.
V2\(\to\)V3 measures the value of \(\delta_m\) calibration; V3\(\to\)V1 the residual
structural cost.
macro = mean of the 7 per-model means (the submission's caliber); micro = mean over
the 27 model\(\times\)benchmark cells, co-reported here (V2: its 21 defined cells
--- V2 makes no prediction on the six non-anchor AIME cells).
V5b answers \S5.3's open question affirmatively on this grid: a base-only predicted
offset performs on par with the RL-calibrated one; its gain over V3 is concentrated
on the 1.5B model.
The lower block traces the full RL-information frontier (appendix, \emph{RL-information
budget frontier}): V5b uses no \emph{target} RL, but its predictor is still fit on the
other six cohort RL models (leave-one-model-out); Z1-N\(^{*}\) drops that to a single
RL anchor cell (and, at 4.04/4.08, actually matches V3 and edges V5b); Z0-u removes RL
entirely, reading \(\alpha\) from base rollouts on the RL-training distribution, at a
cost of only \({\approx}1\) pp. All frontier rungs are defined only on the seven cohort
models (they need base@rltrain), so they do not extend to the new-model block above.}
\end{minipage}
\end{tabular}

\vspace{0.4ex}

\caption{\footnotesize
Extension of Table~\ref{tab:vertical-combined} (vertical transfer and model-offset
analysis; same layout and frozen parameters, no refitting).
(A) adds per-model rows for V3 and the base-only tier V5b, plus a block with two
second-recipe RL models and one cross-family model without calibration data.
(B) adds V5b and co-reports macro and micro means for every tier.
}
\label{tab:reb-vertical}
\end{table}

% ------------------------------------------------------------------ Table R3
\begin{table}[!t]
\scriptsize
\setlength{\tabcolsep}{2pt}
\renewcommand{\arraystretch}{0.96}

\begin{tabular}{@{}p{0.60\textwidth}@{\hspace{\columnsep}}p{0.355\textwidth}@{}}
\begin{minipage}[t]{\linewidth}
\vspace{0pt}
\raggedright
\begin{adjustbox}{max width=\linewidth}
\begin{tabular}{@{}lcccccc@{}}
\toprule
regime map & \(\widehat{N}(b)\) on AIME & Math500 & AIME & GPQA & IFEval & mean \\
\midrule
three regimes (submission): AIME\(\to\)floor, \(\beta\!=\!0\) & 2/2/2/2/2 & 1.12 &
\makecell{2.67\\{\tiny[0.00, 8.33]}} & 6.77 & 3.11 & 3.41 \\
two regimes (counterfactual): AIME\(\to\)math, \(\beta\!=\!0.6\) & 2/4/8/8/16 & 1.12 &
\makecell{1.00\\{\tiny[0.33, 6.67]}} & 6.77 & 3.11 & 3.00 \\
\bottomrule
\end{tabular}
\end{adjustbox}

\panelnote{\textbf{(A)} Regime-map sensitivity on the anchor.
Only the AIME column differs between the two maps (Math500 stays math,
GPQA/IFEval stay nonmath); \(\alpha'{=}2.00\) (floor) vs.\ 2.64 (math).
Paired bootstrap on the AIME cell:
\(\Delta(\text{three}{-}\text{two}) = +1.67\)\,pp, 95\% CI \([-5.00, +4.00]\) ---
contains 0, and the per-budget Wilson intervals of the two paths mutually overlap
(\(n{=}60\)).
The two maps are statistically indistinguishable on this grid, so every headline
number keeps the pre-registered three-regime map (no refitting), ``floor'' is
downgraded to a descriptive label, and the two-regime row is a diagnostic
counterfactual, not an adopted rule.
Values are identical under the H5, V3, and V5b calibers (anchor path identity).}
\end{minipage}
&
\begin{minipage}[t]{\linewidth}
\vspace{0pt}
\raggedright
\begin{adjustbox}{max width=\linewidth}
\begin{tabular}{@{}lccc@{}}
\toprule
AIME, same budget \(b\) & \(b\!=\!1\) & \(b\!=\!16\) & \(b\!=\!256\) \\
\midrule
base envelope (\%) & 3.3 & 11.7 & 25.0 \\
RL default policy (\%) & 6.7 & 10.0 & 20.0 \\
\bottomrule
\end{tabular}
\end{adjustbox}

\panelnote{\textbf{(B)} Same-budget AIME comparison at large budgets (\(n{=}60\),
both sides re-sampled to \(b{=}256\)).
In-sample, the base envelope crosses the RL curve at \(k^{*}{=}8\) and rises
monotonically 11.7\%\(\to\)25.0\% from \(b{=}16\) to 256; Wilson 95\% CIs overlap
at every \(b\) (at \(b{=}256\): base 25.0\% [15.8, 37.2] vs.\ RL 20.0\%
[11.8, 31.8]), so we make no population-level claim about the gap.
Under the frozen floor map the rule keeps \(\widehat{N}\!\equiv\!2\) and its
\(b\!>\!16\) AIME error grows to 12.50\,pp (Table~\ref{tab:reb-horizontal}B): the
large-budget misfit of \(\beta\!=\!0\) is an allocation error of the budget map,
not a capability ceiling of base search --- consistent with the counterfactual
in (A).}
\end{minipage}
\end{tabular}

\vspace{0.4ex}

\caption{\footnotesize
Sensitivity of the \(\beta\!=\!0\) (floor) assignment on the anchor.
(A) The submission's three-regime map vs.\ a two-regime (math/nonmath)
counterfactual that removes the floor class; the difference is inside the AIME
sampling noise.
(B) Same-budget base-vs.-RL evidence on AIME at large budgets.
Frozen parameters throughout; the counterfactual is diagnostic and is not used
for any headline number.
}
\label{tab:reb-beta}
\end{table}

\subsection{Rule variants and robustness checklist}
\label{app:rule-variants}

\subsubsection{Rule variants}

Table~\ref{tab:rule-variants} positions the rule families we considered against the v5 BOPTR-P1 main diagnostic.  The two oracles bound recoverability from above (per-budget oracle) and characterize whether the recovery is coherent across budgets (smooth path).  v5 keeps a low-dimensional state \([P,Q,T]\) and is the rule used throughout the main paper; v6 adds a signed sharpening-vs-expansion axis and helps on compression-dominated cells but hurts the expansion cells, demonstrating that one extra free axis is not free.  v7 fully decouples support and concentration with separate adapters; with a single anchor it under-identifies and worsens transfer everywhere --- this is the negative ablation cited in Section~\ref{sec:boptr-takeaways}.  v8-H is the planned next-step rule family: same low-dim state as v5 but with a small adapter for the model-specific component currently absorbed by \(\delta_m\).

\begin{table}[t]
\centering
\small
\setlength{\tabcolsep}{4pt}
\begin{tabular}{lcccp{0.30\textwidth}}
\toprule
Rule & State & Parameters & Use in paper & Interpretation \\
\midrule
Per-budget oracle & scalar metric & high & upper bound & Shows recoverability but not structure. \\
Smooth path & metric-specific & medium & diagnostic & Tests whether recovery is coherent across budgets. \\
v5 shared-core & [P,Q,T] & low-medium & main diagnostic & Most stable current rule. \\
v6 signed SAT & [S,A,T] & medium & ablation & Demonstrates compression need but hurts expansion. \\
v7 axis-decoupled & [S,A,T] & high & negative ablation & Shows single-anchor under-identification. \\
v8-H planned & [P,Q,T] + adapter & low-medium & future/main candidate & Same rule family with model-specific shifts. \\
\bottomrule
\end{tabular}
\caption{Rule variants considered, positioned against the v5 BOPTR-P1 main diagnostic.  The two oracles bound recoverability from above; v6 and v7 each add one degree of freedom and lose ground where the extra parameter cannot be identified from a single anchor; v8-H is a planned rule family, not evaluated here.}
\label{tab:rule-variants}
\end{table}

\paragraph{Why these six and not more.}
The two oracles (per-budget, smooth-path) are not rules we would deploy --- they refit \(\widehat{\pi}\) or \(\widehat{N}\) per cell --- but they delimit the recoverability ceiling.  Per-budget oracle says ``the support region contains a matching \((\pi,N)\) at every \(b\),'' and smooth-path says ``those matches connect into a budget-coherent trajectory rather than five disjoint lookups.''  v5 (BOPTR-P1) is the smallest rule that respects both constraints from a single calibration cell, which is why we adopt it as the main diagnostic.  v6 and v7 are deliberate ablations: each adds one degree of freedom (signed sharpening axis for v6, independent support/concentration heads for v7) and each loses ground precisely where the extra parameter cannot be re-identified from one anchor.  v8-H is listed as planned, not evaluated --- it replaces the per-model \(\delta_m\) scalar with a small base-feature adapter, which the V4 negative (Section~\ref{sec:boptr-takeaways}) shows requires either more models or rltrain-side features to identify reliably.

\subsubsection{Robustness checklist}
Each item maps a headline claim to the specific analysis cell where an unguarded version would be wrong, together with the guardrail we adopt.
\begin{itemize}[leftmargin=*]
    \item Do not claim parameter-level equivalence between RL and an external decoding policy (Section~\ref{sec:boptr-takeaways}: the rule is a behavioral diagnostic, not a parameter recovery).
    \item Do not claim exact post-RL performance prediction unless a strong checkpoint-selection baseline is included (the rule predicts the operating-point shift, not the absolute pass-rate after training).
    \item Separate horizontal same-model transfer from vertical cross-model transfer (Tables~\ref{tab:regime-horizontal}B and \ref{tab:vertical-combined}A use different baseline families for a reason).
    \item Mark AR-only and expanded-policy pools separately (the recovery near-match search uses both; the BOPTR rule selector uses only AR).
    \item Treat IFEval as evaluation-metric mismatch/OOD rather than as a direct math-style SC benchmark (regime \(=\) nonmath, not math).
    \item Present 1.5B and Llama as stress tests when generation length or protocol behavior differs (the lower-half cohort in Table~\ref{tab:vertical-combined}A is the headline failure case for the V2 direct anchor-path copy).
\end{itemize}

\subsection{Artifact documentation, licenses, and intended use}
\label{app:artifact-documentation}

\paragraph{Artifacts used and citation.}
We use publicly released checkpoints, benchmarks, and software packages.
These include SimpleRL-Zoo checkpoints and scripts, the evaluated model families, Math500/MATH, AIME 2024/2025, GPQA-Diamond, IFEval, vLLM, and HuggingFace/Transformers.
We cite the creators of these artifacts where they are introduced in the main text and appendix.
Our released artifacts consist of code, configuration files, and analysis scripts for reproducing the experiments in this paper.

\paragraph{Licenses, terms, and intended use.}
We use third-party checkpoints, benchmarks, and software packages under their original licenses and terms of use.
All existing artifacts are used only for research and evaluation of model behavior, consistent with their stated intended use and access conditions.
Our released code, configuration files, and analysis scripts are intended for research reproduction and analysis.
They do not redistribute third-party checkpoints or benchmark data.
Users should obtain third-party artifacts from their original sources under their original licenses and terms.

\paragraph{Artifact coverage.}
The evaluation artifacts cover mathematical reasoning, scientific question answering, and instruction following.
Math500/MATH and AIME are English mathematical reasoning benchmarks.
GPQA-Diamond is an English graduate-level science question-answering benchmark.
IFEval is an English instruction-following benchmark with automatically verifiable instructions.
The artifacts are not designed for demographic or social-group analysis, and we do not use them to study demographic-group representation.

\paragraph{Data statistics and splits.}
Math500 contains 500 problems from the MATH benchmark.
AIME 2024 and AIME 2025 contain 30 problems each.
GPQA-Diamond contains 198 multiple-choice science questions.
IFEval contains approximately 500 prompts with verifiable instructions.
We use these public benchmark evaluation sets as provided and do not create new train/dev/test splits.
Our main experiments evaluate budgets \(b\in\{1,2,4,8,16\}\), paired Base/RL checkpoints, and the evaluation metrics described in Section~\ref{sec:setup}.

\paragraph{Personally identifying information and offensive content.}
We do not collect new human-subject data or user-generated personal text.
The benchmarks used in this work are publicly released research/evaluation artifacts for mathematical reasoning, science question answering, and instruction-following evaluation.
We use these benchmarks only through their official evaluation prompts and aggregate metrics, and we do not release raw benchmark contents, model generations tied to individuals, or any derived dataset containing personal identifiers.
We reviewed the benchmark descriptions and task schemas used in our experiments and found no fields designed to collect or expose personal identifiers.
We did not apply additional anonymization because our created artifacts are code, configuration files, and aggregate analysis scripts/results rather than newly collected text data.
Any risk of sensitive or offensive content is inherited from the original public benchmarks, which we cite and use under their stated research/evaluation terms.

\end{document}